\documentclass{article}

\usepackage[main, final]{neurips_2026}

\usepackage{graphicx}
\usepackage[utf8]{inputenc} 
\usepackage[T1]{fontenc}    
\usepackage{hyperref}       
\usepackage{url}            
\usepackage{booktabs}       
\usepackage{amsfonts}       
\usepackage{nicefrac}       
\usepackage{microtype}      

\hypersetup{
	colorlinks=true,
	linkcolor=red,
	filecolor=red,      
	urlcolor=magenta,
	citecolor=blue
}
\usepackage{amsmath}
\usepackage{multicol}
\usepackage{multirow}
\usepackage{placeins}
\usepackage{marvosym}
\usepackage{wrapfig}
\usepackage{algorithm}
\usepackage{courier}
\usepackage{arydshln}
\usepackage[table]{xcolor}

\definecolor{PromptBoxBg}{RGB}{248,248,249}
\definecolor{PromptBoxFrame}{RGB}{210,214,220}
\definecolor{PromptTitleColor}{RGB}{36,92,145}
\definecolor{OutputTitleColor}{RGB}{128,76,36}

\title{
Overcoming Catastrophic Forgetting in Visual Continual Learning with Reinforcement Fine-Tuning
}

\author{
  Meng Lou, Hanzhong Guo, Linwei Chen, Yizhou Yu \\
  School of Computing and Data Science, The University of Hong Kong \\
\small\{\
  \texttt{loumeng@connect.hku.hk},
  \texttt{hanzhong@connect.hku.hk}, \\
  \small
  \texttt{chenlinwei.ai@gmail.com},
  \texttt{yizhouy@acm.org}
\}\ 
  \vspace{-10pt}
}

\begin{document}

\maketitle
\begin{abstract}
Reinforcement Fine-Tuning (RFT) with verifiable rewards has emerged as a powerful learning paradigm for eliciting reasoning capabilities in Multi-modal Large Language Models (MLLMs). 
Recent studies suggest that Reinforcement Fine-Tuning (RFT) is inherently more resilient to catastrophic forgetting than Supervised Fine-Tuning (SFT). However, whether RFT (e.g., GRPO) can effectively overcome forgetting in challenging visual continual learning settings, such as class-incremental learning (CIL) and domain-incremental learning (DIL), remains an open problem. Through a pilot study, we confirm that while RFT consistently outperforms SFT, it still suffers from non-negligible forgetting. We empirically trace this bottleneck to Trajectory-level Drift Agnosticism: among candidate rollouts achieving identical task rewards, the KL divergence from the preceding-task policy varies substantially, which strongly correlates with catastrophic forgetting across sequential tasks. Motivated by this insight, we propose Retention-aware Policy Optimization (RaPO), a simple yet effective RFT method that explicitly mitigates forgetting through trajectory-level reward shaping. Specifically, RaPO comprises two core components: (1) Retention Reward that converts trajectory-level distribution drift into a continuous reward signal, preferentially reinforcing knowledge-preserving rollouts within each group; (2) Cross-Task Advantage Normalization (CTAN), which maintains a persistent exponential moving average of reward statistics across task boundaries to stabilize the optimization progress during continual learning. Leveraging the free-form textual generalization of MLLMs, we comprehensively evaluate RaPO across five visual continual learning settings. Extensive experiments demonstrate that RaPO achieves leading performance, substantially reducing catastrophic forgetting while preserving strong plasticity. To the best of our knowledge, this work represents the first systematic exploration of RFT in visual continual learning, offering insights that we hope will inspire future research.
Code is publicly accessible at: \url{https://github.com/LMMMEng/RaPO}.
\end{abstract}

\vspace{-8pt}
\section{Introduction}
\label{sec:intro}
Reinforcement Fine-Tuning (RFT) with verifiable rewards \cite{team2025kimi15,bai2025qwen3vl,guo2025seed15vl,team2026qwen35,zhang2025rlsurvey,guo2026leveraging} has demonstrated remarkable progress in eliciting reasoning capabilities in Multi-modal Large Language Models (MLLMs), outperforming classical Supervised Fine-Tuning (SFT). GRPO \cite{shao2024deepseekmath} stands as a representative work that leverages verifiable RFT to train powerful large reasoning models \cite{guo2025deepseekr1,liu2025deepseekv32}.
Motivated by this success, several subsequent works \cite{liu2025visualrft,li20252025thinkornot,tan2025reasonrft,he2026finer1} have demonstrated that RFT can effectively improve vision tasks even under few-shot training regimes.
\par
However, real-world applications frequently encounter streaming data within continuously evolving environments \cite{gomes2017survey}, requiring large models to continually adapt to newly arriving data without suffering from catastrophic forgetting \cite{zheng2025towards}.
Although numerous SFT-centered approaches \cite{shi2025continual,yang2025recent,he2026continualinstruction} have been developed for continual learning, recent studies \cite{lai2025rlnaturally,shenfeld2026rl_razor} have revealed that RFT is naturally more resilient to catastrophic forgetting than SFT. 
This property stems from the fact that on-policy learning in RFT implicitly biases the optimization toward solutions residing in low-drift distribution spaces, whereas SFT is prone to converge on solutions within arbitrary distribution drift \cite{lai2025rlnaturally}.
Nevertheless, the efficacy of RFT in challenging visual continual learning, such as class-incremental learning (CIL) \cite{zhou2024cilreview} and domain-incremental learning (DIL) \cite{wang2024clcomprehensive}, remains an open problem.

\begin{figure}[t]
    \centering
    \includegraphics[width=0.85\textwidth]{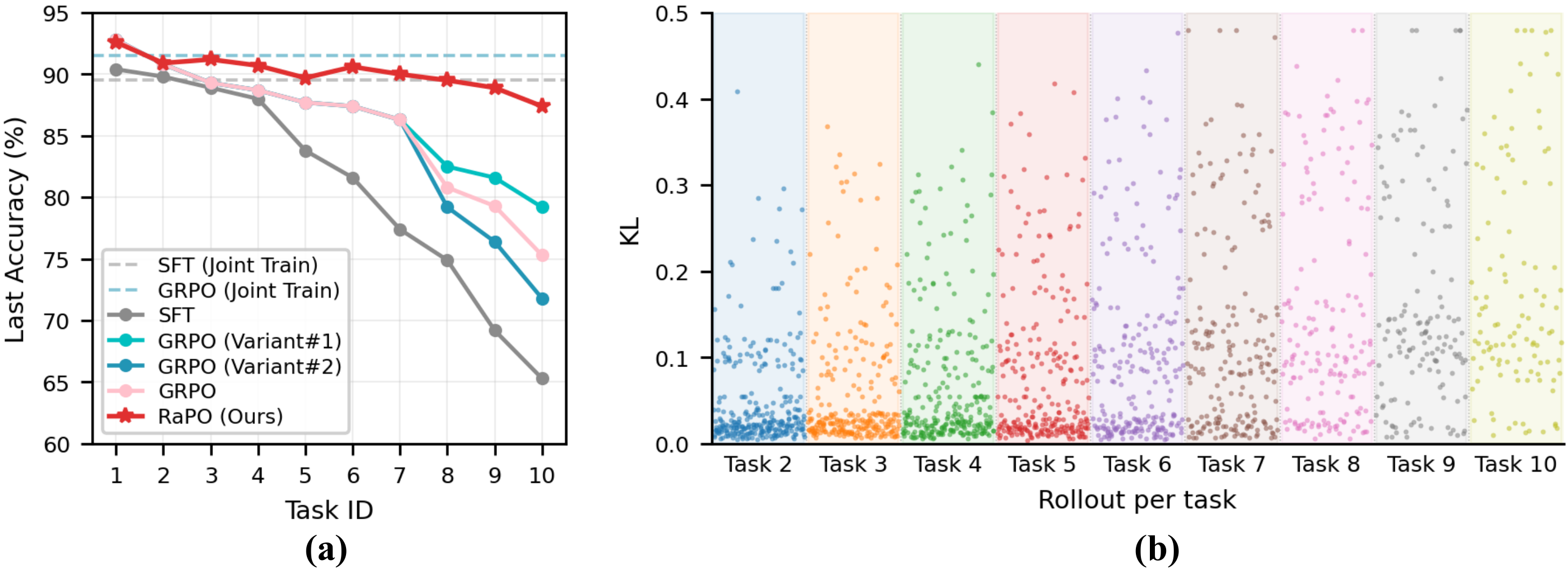}
    \caption{
    \textbf{(a)} RFT clearly outperforms SFT in rehearsal-free CIL, but still suffers from significant forgetting.
    \textbf{(b)} Among equally-rewarded rollouts, KL divergence from the policy in the preceding task varies substantially, and this difference enlarges as tasks progress.
    }
    \label{fig:intro_fig}
    \vspace{-18pt}
\end{figure}

To investigate this, we conduct a pilot study on a challenging rehearsal-free few-shot CIL setting on the widely adopted ImageNet-R dataset~\cite{imagenet-R}. Specifically, 200 image classes are randomly split over 10 non-overlapping tasks, with 20 classes and only 5 labeled examples per class at each task. 
We compare GRPO and SFT based on the Qwen2-VL-2B model~\cite{wang2024qwen2}, together with a joint-training upper bound.
As shown in Figure \ref{fig:intro_fig} (a), GRPO consistently outperforms SFT, confirming that the forgetting resilience of RFT~\cite{lai2025rlnaturally,shenfeld2026rl_razor} successfully transfers to visual tasks.
Nevertheless, \textbf{GRPO still suffers from non-negligible forgetting, demonstrating that it remains insufficient to mitigate stability-plasticity tension in challenging visual continual learning}.
\par
To further explore the rationale of this phenomenon, we analyze the trajectory-level learning patterns of GRPO during CIL.
Specifically, we measure the distribution drift of each rollout generated by the current policy $\pi_t$ as its token-level KL divergence from the frozen preceding-task policy $\pi_{t-1}$.
We focus only on rollout groups with maximal task reward so that the comparison isolates drift differences among equal-reward trajectories. 
As illustrated in Figure \ref{fig:intro_fig} (b), \textbf{candidate rollouts achieving the same task reward exhibit vastly different KL divergence values}, a phenomenon we term \textbf{trajectory-level drift agnosticism}. 
For instance, two distinct trajectories generated from the same input can solve the current task equally well, but exhibit entirely different magnitudes of distributional drift. This discrepancy is increasingly pronounced as the task sequence progresses, which clearly correlates with the accuracy degradation trend shown in Figure \ref{fig:intro_fig} (a). 
This suggests that purely task-reward-driven behavior leads to drift-agnostic credit assignment, which may contribute to severe forgetting.
\par
To validate this hypothesis, we design two simple GRPO variants that intervene on all-correct rollout groups in the later tasks, when forgetting becomes pronounced. Specifically, Variant\#1 assigns zero reward to any rollout whose KL exceeds the group mean, thereby retaining positive reinforcement only for low-drift trajectories. Conversely, Variant\#2 mirrors this operation, preserving positive rewards solely for heavily drifted rollouts. As demonstrated in Figure \ref{fig:intro_fig} (a), these variants exhibit opposed behaviors: Variant\#1 effectively mitigates forgetting, whereas Variant\#2 significantly exacerbates it. 
Collectively, these results validate trajectory-level drift agnosticism as a key empirical phenomenon in vanilla GRPO, i.e., \textbf{trajectory-level drift with respect to the preceding-task policy is an actionable signal closely tied to forgetting}.
However, these hard-thresholding variants are impractical. First, binary gating collapses fine-grained credit assignment required for complex tasks such as dense predictions. Second, it is unstable under small rollout numbers, where even marginal KL differences may trigger winner-take-all updates.
\par
Driven by the above observations, we propose Retention-aware Policy Optimization (RaPO), a simple yet effective RFT method that mitigates catastrophic forgetting through trajectory-level reward shaping. RaPO consists of two complementary components. First, a Retention Reward converts the trajectory-level drift from the preceding-task policy into a dense reward signal: rollouts that stay closer to the preceding policy receive proportionally higher rewards. This design differs fundamentally from standard GRPO, i.e., when two rollouts achieve comparable task rewards but exhibit different degrees of drift, RaPO explicitly reinforces the knowledge-preserving one, steering the policy toward regions that adapt to new data while remaining anchored to previously acquired knowledge. Second, Cross-Task Advantage Normalization (CTAN) maintains a persistent smoother of the reward scale, preventing the abrupt advantage fluctuations that arise from sharp reward-distribution shifts at task boundaries. Together, the Retention Reward directs credit assignment toward low-drift trajectories, while CTAN smoothly stabilizes its scale across the continual learning stream.
\par
Leveraging the free-form textual generalization capabilities of MLLMs, our method has been comprehensively evaluated across a diverse suite of visual continual learning tasks, including class-incremental image classification, domain-incremental image classification, class-incremental object detection, domain-incremental object detection, and class-incremental video classification. Extensive experimental results in Section \ref{sec:experiments} demonstrate the promising performance and generalization capacity of RaPO.
Overall, our goal is not to chase state-of-the-art performance on different benchmarks, but to systematically explore the potential of verifier-based RFT for visual continual learning. We hope this work will stimulate further research on RFT-based continual learning.

\vspace{-13pt}
\section{Related Work}
\label{sec:related_work}
\vspace{-8pt}
\textbf{Reinforcement Fine-Tuning (RFT)} has demonstrated a superior capacity to incentivize reasoning capabilities in LLMs \cite{jaech2024openaio1}. A foundational paradigm is reinforcement learning with human feedback, which aligns model outputs with human preferences \cite{schulman2017proximal, ouyang2022training, rafailov2023direct, dai2024safe}. Recently, the research has increasingly shifted toward reinforcement learning with verifiable rewards. In particular, the remarkable success of Group Relative Policy Optimization (GRPO) \cite{shao2024deepseekmath, guo2025deepseekr1} has motivated a surge of research exploring this paradigm further, such as Dr.GRPO \cite{liu2025drgrpo}, DAPO \cite{yu2025dapo}, and GSPO \cite{zheng2025gspo}.
Subsequently, many works \cite{liu2025visualrft, tan2025reasonrft, he2026finer1, feng2025onethinker} have also demonstrated that RFT can significantly improve visual tasks over SFT by activating reasoning capabilities. 
Additionally, recent studies \cite{lai2025rlnaturally,shenfeld2026rl_razor} suggest that RFT is naturally more resistant to catastrophic forgetting than SFT when adapting to new data. 
\par
\textbf{Visual Continual Learning}.
Recent years have witnessed rapid progress in diverse visual tasks \cite{lou2025overlock,lou2025sparx,lou2025transxnet,fu2025segman,chen2022compound,dai2026cgsa,dai2026oco,dai2026cpf}.
However, visual continual learning remains a long-standing problem that aims to adapt models to non-stationary visual streams without catastrophic forgetting \cite{wang2024clcomprehensive}. 
Among its various paradigms, rehearsal-free class-incremental learning (CIL) \cite{wang2022l2p,zhou2025aper} stands out as one of the most representative and challenging settings, which requires a model to continuously adapt to incrementally arriving classes without accessing any historical training data, while simultaneously preserving its recognition capabilities on all observed classes. During CIL, the label spaces across different tasks are strictly disjoint. 
Another prominent setting is rehearsal-free domain-incremental learning (DIL) \cite{wang2022sprompt,wang2024non}, which aims to enable a model to sequentially adapt to new domains while ensuring its previously acquired knowledge is not catastrophically degraded by domain shifts. Unlike CIL, different tasks in DIL share the same label space but exhibit distinct domain distributions.
Existing progress \cite{zhou2024continualptm,lou2026scaling,fu2026clare,fu2026one} has been driven primarily by SFT-based adaptation of vision-centric models. 
One of the most prevalent paradigms involves incrementally appending parameter-efficient modules \cite{lou2026parameters,liang2024inflora,yu2024moe_adapters,wang2025tuna,sun2025mos,zhou2025dualcon} into pre-trained models such as ViT \cite{dosovitskiy2020vit} and CLIP \cite{radford2021clip}, demonstrating promising results.
Nevertheless, these methods are typically centered on a single scenario, such as class-incremental image classification, rather than a unified model that is able to handle diverse settings simultaneously, including class-incremental image/video classification and dense predictions.
On the other hand, since real-world scenarios frequently lack abundant, high-quality annotations for each incremental task, vision models with SFT may be prone to overfitting under data-scarce conditions.
\par
In this work, we explore the untapped potential of RFT for challenging visual continual learning paradigms. 
Without bells and whistles, our proposed RaPO achieves leading performance and strong generalizability compared with different baselines.
To the best of our knowledge, this work is the first to systematically explore the potential of RFT in visual continual learning.

\vspace{-9pt}
\section{Method}
\label{sec:method}
\vspace{-8pt}
\subsection{Preliminaries}
\label{sec:prelim}
\vspace{-7pt}
\textbf{GRPO} \cite{shao2024deepseekmath} optimizes a policy $\pi_\theta$ using group-relative advantages instead of a learned value function. For a given textual prompt $x$, GRPO samples a rollout group $\mathcal{G}(x)=\{y_1,\ldots,y_n\}$ of $n$ trajectories from the current policy $\pi_\theta$. Each rollout $y_i$ receives a task-specific verifiable reward $R_{\mathrm{task}}(y_i)$. These rewards are centered and rescaled within the group to produce the group-relative advantage $A_i$:
\begin{equation}
A_i \; = \; \frac{R_{\mathrm{task}}(y_i) - \mu_{\mathrm{group}}}{\sigma_{\mathrm{group}} + \epsilon}
\label{eq:grpo_advantage}
\end{equation}
where $\mu_{\mathrm{group}}$ and $\sigma_{\mathrm{group}}$ are the mean and standard deviation of the rewards within $\mathcal{G}(x)$, respectively. Then, the policy is updated using the clipped surrogate objective \cite{schulman2017proximal}, where the standard advantage based on the value-function is replaced by the group-relative advantage $A_i$.
\par
\textbf{RFT in Vision.} 
To address visual continual learning, the input is defined as a multi-modal signal $x=(v, l)$, where $v$ is a visual input (image or video) and $l$ is a textual instruction template specifying the task. The output $y_i$ is a textual response generated by MLLM. Following the recent paradigm \cite{liu2025visualrft}, we employ diverse verifiable reward functions $R_{\mathrm{task}}$ depending on the task type, such as accuracy reward for image/video classification and IoU reward for object detection. 
More details concerning prompt templates and reward formulations are provided in Appendix~\ref{sec:append_implementations}.

\vspace{-5pt}
\subsection{Retention-aware Policy Optimization}
\label{sec:rapo}
\vspace{-5pt}
\subsubsection{Overview}
\vspace{-5pt}
We study rehearsal-free visual continual learning (i.e., CIL and DIL) across a sequential stream of tasks $\{\mathcal{T}_1, \ldots, \mathcal{T}_N\}$. For an arriving task $\mathcal{T}_t$, the learner is optimized using only the training data from $\mathcal{T}_t$, with no access to historical training data from $\mathcal{T}_1$ to $\mathcal{T}_{t-1}$. The overall learning pipeline is simple. Specifically, at the onset of task $\mathcal{T}_t$, we initialize the actor policy $\pi_t$ using the weights saved at the end of $\mathcal{T}_{t-1}$, while simultaneously maintaining a frozen copy of it to serve as the anchor policy $\pi_{t-1}$. 
During each optimization iteration, the actor $\pi_t$ generates a group of candidate rollouts for a given multi-modal input $x=(v, l)$. Each rollout is then evaluated across two aspects: 
1) The primary task reward is computed by a task-specific verifier. 
2) A new retention reward is estimated with a trajectory-level drift metric against the anchor $\pi_{t-1}$.
These two signals are aggregated into a unified objective to update $\pi_t$, explicitly steering the policy to explore parameter spaces that jointly maximize proficiency in new tasks and historical knowledge preservation. Concurrently, to counter the optimization instability caused by abrupt reward-distribution shifts when transitioning across task boundaries, a simple Cross-Task Advantage Normalization (CTAN) mechanism is introduced to regulate the scale of the credit assignment.

\subsubsection{Retention Reward}
\label{sec:rapo_ret}
\vspace{-5pt}
As empirically validated in the Section \ref{sec:intro}, vanilla GRPO suffers from trajectory-level drift agnosticism. When the actor policy $\pi_t$ adapts to a new task, the trajectory-level distribution drift from the anchor policy $\pi_{t-1}$ is strongly correlated with the catastrophic forgetting of previously acquired knowledge. This inspires us to explicitly formulate this trajectory-level drift into a continuous reward signal.
Specifically, let $y_i = (y_{i,1}, \ldots, y_{i,m_i})$ denote a rollout sample of length $m_i$ sampled from $\pi_t$, where $i \in \{1,\ldots,n\}$ indexes the rollouts within the group and $s \in \{1,\ldots,m_i\}$ indexes the token positions. 
Suppose $y_{i,<s}=(y_{i,1},\ldots,y_{i,s-1})$ represent the generated prefix up to step $s$, the trajectory-level distribution drift is calculated as:
\begin{equation}
\bar{D}_{\mathrm{drift}}(y_i) \;=\; \max\! \; ( \frac{1}{m_i}\sum_{s=1}^{m_i} \left[\log \pi_t(y_{i,s}\mid x, y_{i,<s}) - \log \pi_{t-1}(y_{i,s}\mid x, y_{i,<s})\right], 0\, )
\label{eq:rapo_dbar}
\end{equation}
There are two remarks on Equation~\eqref{eq:rapo_dbar} that warrant highlighting. 
First, the bracketed term computes the per-token log-probability ratio between the actor $\pi_t$ and the anchor $\pi_{t-1}$. 
Averaging this ratio over the $m_i$ generated tokens provides a length-normalized Monte Carlo estimate of the trajectory-level distribution drift evaluated along the sampled trajectory.
This length normalization is crucial, as it ensures that $\bar{D}_{\mathrm{drift}}(y_i)$ remains strictly comparable across candidate rollouts of varying lengths. 
Second, the outer $\max(\cdot, 0)$ applies a one-sided truncation. Specifically, a negative pre-truncation value indicates that the actor $\pi_t$ has become less confident in the generated trajectory than the anchor $\pi_{t-1}$, signifying that $\pi_t$ has not specialized toward this trajectory relative to $\pi_{t-1}$. Therefore, clamping these negative values to zero explicitly prevents reward hacking in the subsequent reward formulation (Equation~\eqref{eq:rapo_ret}), as the actor $\pi_t$ is possible to intentionally generate low-confidence outputs to inflate its reward.
\par
To seamlessly incorporate this forgetting measurement into the RFT objective, we stop the gradient of $\bar{D}_{\mathrm{drift}}(y_i)$ and convert it into a bounded and positive reward score using an exponentially decaying mapping:
\begin{equation}
R_{\mathrm{ret}}(y_i) \;=\; \exp\!\left(-\alpha \, \bar{D}_{\mathrm{drift}}(y_i)\right) \in (0, 1]
\label{eq:rapo_ret}
\end{equation}
where $\alpha > 0$ is a scaling hyperparameter that controls the sensitivity to distribution drift. Rollouts that remain closer to the anchor $\pi_{t-1}$ yield a lower $\bar{D}_{\mathrm{drift}}$, thereby translating into a higher $R_{\mathrm{ret}}$ score approaching $1.0$. Since this reward directly measures the retention of previously learned knowledge, it is termed the retention reward. 
Afterwards, the retention reward is seamlessly integrated with the task-specific reward via an additive formulation:
\begin{equation}
R_{\mathrm{total}}(y_i) \;=\; R_{\mathrm{task}}(y_i) + \lambda\, R_{\mathrm{ret}}(y_i)
\label{eq:rapo_total_reward}
\end{equation}
where $\lambda > 0$ balances task adaptation and retention. Crucially, $R_{\mathrm{ret}}$ enters the composite reward before the group-relative advantage computation. This explicitly changes the rollout ranking inside $\mathcal{G}(x)$: among candidate trajectories that achieve comparable task rewards on $\mathcal{T}_t$, those remaining closer to the anchor $\pi_{t-1}$ are assigned larger advantages.
\par
Despite its simplicity, this formulation provides three properties: 
\textit{Continuous:} within a similar-$R_{\mathrm{task}}$ sub-group, the assigned advantage is an increasing function of the $R_{\mathrm{ret}}$, ensuring that the anchor-closer rollout is always reinforced more strongly.
\textit{Disentangled:} The weighted-additive form limits the influence of retention relative to task reward. Since $R_{\mathrm{ret}}$ is strictly bounded, the retention term can change the total reward by at most $\lambda$, so its effect remains controlled and is most relevant when candidate trajectories have similar task rewards.
\textit{General:} The retention reward can be easily combined with any task reward, such as a binary accuracy reward for image classification and a continuous IoU reward for object detection. 
Note that retention reward differs fundamentally from standard loss-level KL regularization, as discussed in Section \ref{sec:method_discussion}.

\subsubsection{Cross-Task Advantage Normalization}
\label{sec:rapo_can}
\vspace{-5pt}
During continual learning, abrupt changes in reward statistics may occur at task transitions. Specifically, near the end of $\mathcal{T}_{t-1}$, many rollouts are similarly correct and the within-batch reward spread $\sigma_{\mathrm{batch}}$ becomes small, which inflates normalized advantages. 
At the start of $\mathcal{T}_t$, rewards typically become lower and more variable, compressing advantages precisely when fast adaptation is required. This oscillation destabilizes GRPO training across task boundaries.
\par
\begin{figure}[t]
    \centering
    \includegraphics[width=0.875\textwidth]{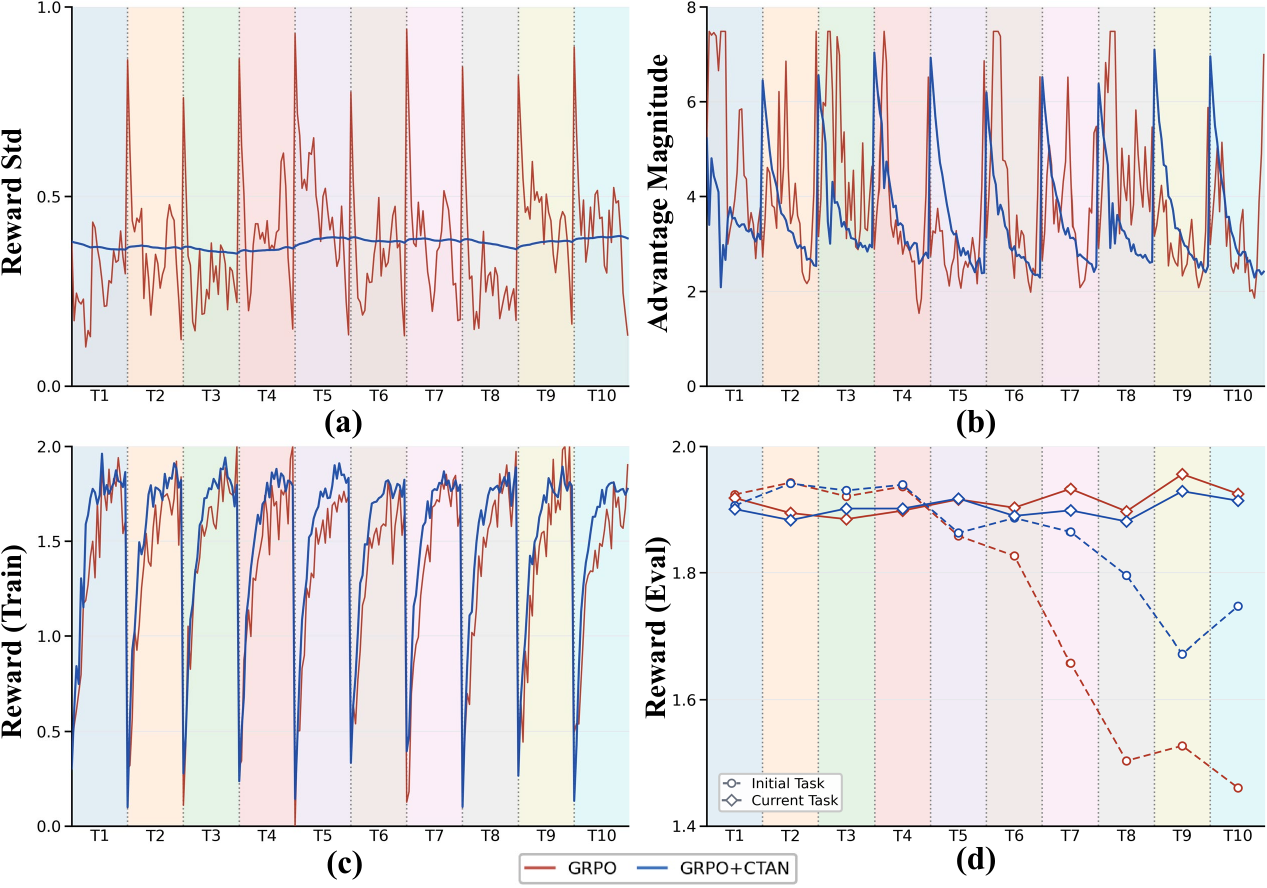}
\caption{
\textbf{(a)} GRPO relies on the instantaneous reward standard deviation, which fluctuates sharply across task boundaries, whereas CTAN maintains a persistent EMA normalizer ($\beta$=0.99).
\textbf{(b)} CTAN produces a smoother advantage magnitude (sum of the absolute values of advantages) across the continual learning stream.
\textbf{(c)} The stabilized advantage scale is accompanied by smoother acquisition of training reward.
\textbf{(d)} Evaluation reward trends further reflect more robust continual learning behavior, with Initial Task denoting evaluation on task\#$1$ and Current Task denoting evaluation on the task currently being learned.
\textit{Overall, CTAN stabilizes cross-task reward normalization, leading to smoother credit assignment, steadier reward acquisition, and stronger continual learning capability.}
}
\label{fig:ema}
\vspace{-23pt}
\end{figure}

To stabilize the optimization scale, we replace the per-batch reward standard deviation with a running exponential moving average (EMA) $\hat{\sigma}$ that is persistent across optimization steps and across task boundaries. At every optimization step, after computing the batch-level reward standard deviation $\sigma_{\mathrm{batch}}$ from the current batch of rollout rewards, we perform the update:
\begin{equation}
\hat{\sigma} \;\leftarrow\; \beta \, \hat{\sigma} \;+\; (1-\beta)\, \sigma_{\mathrm{batch}}
\label{eq:tan_update}
\end{equation}
where $\beta$ is the smoothing coefficient (e.g., $\beta=0.99$). We then compute the advantage as:
\begin{equation}
A_i \;=\; \frac{R_{\mathrm{total}}(y_i) - \mu_{\mathrm{group}}}{\hat{\sigma} + \epsilon}
\label{eq:tan_advantage}
\end{equation}
where $\mu_{\mathrm{group}}$ is still the mean total reward within the rollout group for the same prompt. The numerator preserves within-group ranking, while the denominator is stabilized across batches and across tasks. CTAN is persistent at task boundaries: the state $\hat{\sigma}$ is saved at the end of $\mathcal{T}_{t-1}$ and loaded as the initial EMA value at the beginning of $\mathcal{T}_t$, so the normalization scale does not reset abruptly when a new task arrives. 
As shown in Figure~\ref{fig:ema}, CTAN exhibits a smoother advantage scale and steadier reward acquisition during continual learning by stabilizing the reward scale across task boundaries.
\par
We have provided more in-depth analysis regarding optimization properties of RaPO in Appendix~\ref{sec:appendix_optimization_stability}. Briefly, the retention reward is compatible with the standard score-function policy-gradient update when treated as detached scalar feedback. The resulting reward and CTAN-normalized advantage remain bounded. Under the usual smoothness, bounded-variance, and score-function moment assumptions, the idealized detached surrogate inherits the standard stationary-point convergence behavior of stochastic policy-gradient methods.

\vspace{-5pt}
\subsection{Discussions of RaPO}
\label{sec:method_discussion}
\vspace{-5pt}
\textbf{Retention Reward vs. Standard KL Loss.}
Although both our retention reward and the KL loss exploit a divergence signal, they intervene at different stages of optimization. The standard loss-level KL term in GRPO, $\mathrm{KL}(\pi_t\,\|\,\pi_{\mathrm{ref}})$, is a regularizer added to the training objective $\mathcal{L}$ (with $\pi_{\mathrm{ref}}$ a fixed reference model) that globally constrains the magnitude of the policy update.
However, its gradient contribution is independent of the task reward received by each rollout, and it cannot distinguish the degree of forgetting within a group of rollouts that share comparable task rewards. In contrast, our retention reward is injected before the group-relative advantage computation, i.e., it lives inside $R_{\mathrm{total}}$ rather than inside $\mathcal{L}$. Consequently, RaPO can change the ranking of rollouts within a group, whereas a standard KL loss term primarily controls the scale of policy movement and cannot differentiate rollouts that receive the same reward. Experimental results in Appendix~\ref{sec:appendix_ablation_kl} confirm that the standard loss-level KL has only a modest effect on continual learning performance.
\par
\textbf{Relation to On-policy Distillation.}
On-policy Distillation (OPD)~\cite{agarwal2024onpolicydist,yang2026learning} trains a student policy on its own sampled trajectories using dense token-level KL supervision from a superior external teacher. This mechanism is inherently tailored to a single-task knowledge transfer paradigm: every generated token contributes an independent gradient that uniformly pulls the student's per-token distribution toward that of the teacher. 
In visual continual learning, however, none of these premises hold. At task $\mathcal{T}_t$, there is no stronger teacher for the incoming data, while the only available reference is the policy in the preceding task (i.e., $\pi_{t-1}$). Consequently, enforcing a dense, token-level KL penalty against $\pi_{t-1}$ on the data of $\mathcal{T}_t$ can inject an independent gradient term that actively pushes $\pi_t$'s per-token distribution back toward $\pi_{t-1}$ at every position. This suppresses the plasticity required to learn $\mathcal{T}_t$, transforming a retention mechanism into an obstacle for adaptation to new tasks.

\vspace{-15pt}
\section{Experiments}
\label{sec:experiments}
\vspace{-15pt}
We conduct a comprehensive experimental evaluation on challenging visual continual learning problems across different visual modalities and task formulations. Specifically, we focus on class-incremental image classification, domain-incremental image classification, class-incremental video classification, class-incremental object detection, and domain-incremental object detection. 
The default hyperparameters of RaPO are $n=8$ (Section \ref{sec:prelim}), $\alpha=20$ (Section \ref{sec:rapo_ret}), $\lambda=0.5$ (Section \ref{sec:rapo_ret}), and $\beta=0.999$ (Section \ref{sec:rapo_can}).
The retention reward is activated starting from task 2, since the pre-trained base model lacks domain-specific knowledge, while anchor $\pi_1$ is obtained only after the model has been adapted to task 1.
All experiments are conducted on 8 NVIDIA H100 GPUs. Due to page limits, more experimental results are provided in Appendix \ref{sec:more_ablation}.

\begin{table}[t]
  \centering
  \caption{
  A comparison of different methods on class-incremental image classification.
  }
  \vspace{-0.5em}
  \resizebox{0.875\textwidth}{!}{
    \begin{tabular}{l cccccccc}
    \toprule
    \multirow{3}[2]{*}{\textbf{Method}} & \multicolumn{4}{c}{\textbf{ImageNet-R}} & \multicolumn{4}{c}{\textbf{ImageNet-A}} \\
    \cmidrule(lr){2-5} \cmidrule(lr){6-9}
    & \multicolumn{2}{c}{\textbf{10 Tasks}} & \multicolumn{2}{c}{\textbf{20 Tasks}} & \multicolumn{2}{c}{\textbf{10 Tasks}} & \multicolumn{2}{c}{\textbf{20 Tasks}} \\
    \cmidrule(lr){2-3} \cmidrule(lr){4-5} \cmidrule(lr){6-7} \cmidrule(lr){8-9}
    & $\mathcal{A}\uparrow$ & $\mathcal{F}\downarrow$ & $\mathcal{A}\uparrow$ & $\mathcal{F}\downarrow$ & $\mathcal{A}\uparrow$ & $\mathcal{F}\downarrow$ & $\mathcal{A}\uparrow$ & $\mathcal{F}\downarrow$ \\
    \midrule
    \multicolumn{9}{c}{\textbf{Joint Training}} \\
    SFT & 89.55 & - & 89.55 & - & 67.61 & - & 67.61 & - \\
    GRPO & 91.50 & - & 91.50 & - & 69.81 & - & 69.81 & - \\
    \midrule
    \multicolumn{9}{c}{\textbf{Continual Learning}} \\
    SFT & 64.49\textsubscript{$\pm$2.44} & 39.83\textsubscript{$\pm$4.99} & 57.84\textsubscript{$\pm$4.87} & 41.91\textsubscript{$\pm$4.96} & 28.18\textsubscript{$\pm$4.97} & 66.92\textsubscript{$\pm$2.80} & 14.20\textsubscript{$\pm$5.99} & 79.71\textsubscript{$\pm$6.78} \\
    L2      & 60.99\textsubscript{$\pm$3.50} & 39.39\textsubscript{$\pm$4.59} & 54.09\textsubscript{$\pm$4.05} & 45.74\textsubscript{$\pm$7.38} & 28.22\textsubscript{$\pm$2.57} & 66.97\textsubscript{$\pm$3.65} & 12.16\textsubscript{$\pm$4.86} & 82.30\textsubscript{$\pm$5.43} \\
    EWC     & 62.47\textsubscript{$\pm$2.69} & 37.75\textsubscript{$\pm$4.65} & 56.97\textsubscript{$\pm$3.49} & 42.74\textsubscript{$\pm$3.67} & 30.37\textsubscript{$\pm$6.56} & 64.24\textsubscript{$\pm$2.96} & 12.64\textsubscript{$\pm$3.43} & 81.75\textsubscript{$\pm$4.21} \\
    LwF     & 69.81\textsubscript{$\pm$1.63} & 26.82\textsubscript{$\pm$2.36} & 57.83\textsubscript{$\pm$2.97} & 40.40\textsubscript{$\pm$6.81} & 23.00\textsubscript{$\pm$3.18} & 68.53\textsubscript{$\pm$3.25} & 11.39\textsubscript{$\pm$3.56} & 81.85\textsubscript{$\pm$3.69} \\
    GRPO    & 74.67\textsubscript{$\pm$1.27} & 20.02\textsubscript{$\pm$4.91} & 73.18\textsubscript{$\pm$3.32} & 17.40\textsubscript{$\pm$5.04} & 37.37\textsubscript{$\pm$2.68} & 28.88\textsubscript{$\pm$4.43} & 22.80\textsubscript{$\pm$3.21} & 48.14\textsubscript{$\pm$5.37} \\
    \rowcolor[rgb]{ 0.8, 1, 1}\textbf{RaPO (Ours)} & \textbf{85.92}\textsubscript{$\pm$1.82} & \textbf{4.69}\textsubscript{$\pm$1.71} & \textbf{81.97}\textsubscript{$\pm$2.88} & \textbf{9.54}\textsubscript{$\pm$3.44} & \textbf{44.61}\textsubscript{$\pm$2.36} & \textbf{20.16}\textsubscript{$\pm$4.62} & \textbf{32.15}\textsubscript{$\pm$0.92} & \textbf{35.19}\textsubscript{$\pm$2.09} \\
    \midrule
    \midrule
    \multirow{3}[2]{*}{\textbf{Method}} & \multicolumn{4}{c}{\textbf{TinyImageNet}} & \multicolumn{4}{c}{\textbf{CUB-200}} \\
    \cmidrule(lr){2-5} \cmidrule(lr){6-9}
    & \multicolumn{2}{c}{\textbf{10 Tasks}} & \multicolumn{2}{c}{\textbf{20 Tasks}} & \multicolumn{2}{c}{\textbf{10 Tasks}} & \multicolumn{2}{c}{\textbf{20 Tasks}} \\
    \cmidrule(lr){2-3} \cmidrule(lr){4-5} \cmidrule(lr){6-7} \cmidrule(lr){8-9}
    & $\mathcal{A}\uparrow$ & $\mathcal{F}\downarrow$ & $\mathcal{A}\uparrow$ & $\mathcal{F}\downarrow$ & $\mathcal{A}\uparrow$ & $\mathcal{F}\downarrow$ & $\mathcal{A}\uparrow$ & $\mathcal{F}\downarrow$ \\
    \midrule
    \multicolumn{9}{c}{\textbf{Joint Training}} \\
    SFT & 34.71 & - & 34.71 & - & 48.79 & - & 48.79 & - \\
    GRPO & 71.51 & - & 71.51 & - & 52.16 & - & 52.16 & - \\
    \midrule
    \multicolumn{9}{c}{\textbf{Continual Learning}} \\
    SFT & 15.94\textsubscript{$\pm$5.93} & 43.29\textsubscript{$\pm$6.74} & 8.71\textsubscript{$\pm$1.82}  & 53.68\textsubscript{$\pm$5.15} & 19.09\textsubscript{$\pm$2.72} & 71.29\textsubscript{$\pm$2.07} & 15.78\textsubscript{$\pm$3.55} & 72.46\textsubscript{$\pm$3.19} \\
    L2      & 15.18\textsubscript{$\pm$3.83} & 44.10\textsubscript{$\pm$4.22} & 8.60\textsubscript{$\pm$3.26}  & 53.98\textsubscript{$\pm$4.51} & 18.39\textsubscript{$\pm$3.56} & 71.56\textsubscript{$\pm$2.63} & 16.70\textsubscript{$\pm$2.66} & 71.72\textsubscript{$\pm$3.48} \\
    EWC     & 17.17\textsubscript{$\pm$6.40} & 41.55\textsubscript{$\pm$6.77} & 11.41\textsubscript{$\pm$0.29} & 51.02\textsubscript{$\pm$0.95} & 18.40\textsubscript{$\pm$2.80} & 71.73\textsubscript{$\pm$1.69} & 14.15\textsubscript{$\pm$4.08} & 74.38\textsubscript{$\pm$3.29} \\
    LwF     & 15.07\textsubscript{$\pm$3.13} & 36.41\textsubscript{$\pm$4.10} & 12.51\textsubscript{$\pm$3.40} & 45.82\textsubscript{$\pm$3.26} & 23.05\textsubscript{$\pm$2.22} & 54.49\textsubscript{$\pm$2.70} & 12.28\textsubscript{$\pm$2.10} & 71.80\textsubscript{$\pm$1.12} \\
    GRPO    & 46.84\textsubscript{$\pm$2.80} & 36.24\textsubscript{$\pm$3.74} & 42.08\textsubscript{$\pm$4.49} & 40.85\textsubscript{$\pm$5.25} & 28.67\textsubscript{$\pm$1.54} & 45.50\textsubscript{$\pm$2.15} & 24.67\textsubscript{$\pm$4.59} & 47.04\textsubscript{$\pm$1.79} \\
    \rowcolor[rgb]{ 0.8, 1, 1}\textbf{RaPO (Ours)} & \textbf{62.36}\textsubscript{$\pm$2.26} & \textbf{16.94}\textsubscript{$\pm$3.94} & \textbf{62.20}\textsubscript{$\pm$3.63} & \textbf{14.68}\textsubscript{$\pm$4.56} & \textbf{45.15}\textsubscript{$\pm$1.20} & \textbf{19.10}\textsubscript{$\pm$1.79} & \textbf{37.33}\textsubscript{$\pm$2.05} & \textbf{25.40}\textsubscript{$\pm$1.13} \\
    \bottomrule
    \end{tabular}%
  }
\label{tab:cls_cl}
\vspace{-15pt}
\end{table}

\vspace{-8pt}
\subsection{Class-Incremental Image Classification}
\label{sec:cil_cls}
\vspace{-8pt}
We first evaluate RaPO under class-incremental image classification, a representative and challenging continual learning setting. The model receives a sequence of tasks, each introducing a disjoint subset of novel classes without access to previous training data. At test time, the model is evaluated on all classes observed so far.
\par
\textbf{Datasets.}
Experiments are presented on four commonly used image classification datasets: ImageNet-R~\cite{imagenet-R}, ImageNet-A~\cite{imagenet-A}, TinyImageNet~\cite{le2015tinyimagenet}, and CUB-200~\cite{wah2011cub200}. For each dataset, we employ two task settings: 10 tasks with 20 classes per task and 20 tasks with 10 classes per task. All results are averaged over three random class orders to reduce the influence of any particular ordering. In practice, we adopt a 5-shot training protocol: only five labeled samples per class are available at each task. 
This scarce-annotation regime mirrors the constraints of real-world deployment, which lacks large and high-quality per-task annotation budgets, and poses a technical challenge that demands efficient few-shot adaptation while retaining prior knowledge  \cite{zhang2025fewcilreview}.
This is also the standard setup in recent vision-centric RFT works~\cite{liu2025visualrft,tan2025reasonrft,he2026finer1}, where learning under limited annotations is central to the problem formulation.
\par
\textbf{Evaluation Metrics.}
Two standard metrics for continual learning \cite{zhou2024cilreview,wang2024clcomprehensive,chaudhry2018riemannian} are reported. Specifically, Last Accuracy ($\mathcal{A}$) is defined as the accuracy over the test set of all observed classes after the model finishes learning the final task. This metric evaluates the overall recognition ability retained by the final model. Forgetting ($\mathcal{F}$) measures the average performance drop from the historical best accuracy on previous tasks to their final accuracy after learning all tasks, which reflects the degree of forgetting throughout the continual learning process.
\begin{wraptable}{r}{0.495\textwidth}
\centering
\vspace{-14pt}
\caption{A comparison of different methods on class-incremental object detection using the COCO 2017 dataset.}
\label{tab:det_cl}
\resizebox{\linewidth}{!}{%
  \begin{tabular}{l cccc}
  \toprule
  \multirow{2}[2]{*}{\textbf{Method}} & \multicolumn{2}{c}{\textbf{5 Tasks}} & \multicolumn{2}{c}{\textbf{10 Tasks}} \\
  \cmidrule(lr){2-3} \cmidrule(lr){4-5} 
  & $\mathcal{A}_b\uparrow$ & $\mathcal{F}_b\downarrow$ & $\mathcal{A}_b\uparrow$ & $\mathcal{F}_b\downarrow$ \\
  \midrule
  \multicolumn{5}{c}{\textbf{Joint Training}} \\
  SFT & 20.39 & - & 20.39 & - \\
  GRPO & 23.48 & - & 23.48 & - \\
  \midrule
  \multicolumn{5}{c}{\textbf{Continual Learning}} \\
  SFT     & 7.09\textsubscript{$\pm$0.28} & 13.37\textsubscript{$\pm$1.04} & 7.41\textsubscript{$\pm$3.37} & 12.51\textsubscript{$\pm$4.36} \\
  L2      & 7.64\textsubscript{$\pm$0.56} & 14.24\textsubscript{$\pm$1.50} & 7.82\textsubscript{$\pm$3.95} & 13.72\textsubscript{$\pm$4.67} \\
  EWC     & 7.14\textsubscript{$\pm$0.52} & 14.51\textsubscript{$\pm$0.43} & 7.87\textsubscript{$\pm$4.01} & 13.91\textsubscript{$\pm$4.79} \\
  LwF     & 13.93\textsubscript{$\pm$0.50} & 7.31\textsubscript{$\pm$0.07}  & 13.41\textsubscript{$\pm$2.25} & 7.34\textsubscript{$\pm$1.91} \\
  GRPO    & 14.64\textsubscript{$\pm$2.18} & 6.67\textsubscript{$\pm$2.71}  & 14.30\textsubscript{$\pm$3.02} & 6.73\textsubscript{$\pm$3.51} \\
  \rowcolor[rgb]{ 0.8, 1, 1}\textbf{RaPO (Ours)} & \textbf{19.31}\textsubscript{$\pm$1.08} & \textbf{1.39}\textsubscript{$\pm$1.20} & \textbf{19.12}\textsubscript{$\pm$0.09} & \textbf{1.37}\textsubscript{$\pm$0.56} \\
  \bottomrule
  \end{tabular}%
}
\vspace{-10pt}
\end{wraptable}
\textbf{Baselines.}
We employ Qwen2-VL-2B \cite{wang2024qwen2} as the base MLLM. This choice is motivated by two considerations. First, stronger frontier MLLMs (e.g., Qwen3 series \cite{team2026qwen35,yang2025qwen3}) may already possess sufficient recognition ability on these benchmarks, making the gains from continual learning difficult to observe. 
Second, image classification is a relatively simple visual task. A relatively small-scale model is sufficiently expressive for continual image classification, while larger models risk overfitting.
We have conducted experiments on a diverse set of baselines. As an upper bound, we jointly train the model on the entire dataset using both SFT and GRPO. For standard continual training, we sequentially apply SFT and GRPO, i.e., each method is used to train the model task by task. We also compare three representative SFT-based baselines tailored for continual learning: 1) A simple L2 regularizer pulls the current policy parameters toward the previous‑task policy during training on each new task; 2) EWC~\cite{kirkpatrick2017ewc} slows down updates on parameters that are important for previous tasks with Fisher information matrices; 3) LwF~\cite{li2017l2f} uses the previous‑task model as a teacher and distills its output distributions on the current task data into the current model. In practice, we train all models for 2 epochs, since more epochs do not improve performance and may lead to overfitting. Note that our goal is not to exhaustively benchmark existing continual learning algorithms, many of which rely on architecture- or task-specific mechanisms, but to isolate whether RFT can serve as a visual continual learning paradigm and whether our design further improves it. We therefore compare representative SFT-based continual-learning baselines and a GRPO-based RFT baseline.

\textbf{Results.}
Table~\ref{tab:cls_cl} summarizes the results on four class-incremental image classification benchmarks. RaPO consistently achieves the best performance across all datasets and task splits. For example, on ImageNet-R with 10 tasks, RaPO improves accuracy from 74.67\% to 85.92\% and reduces forgetting from 20.02\% to 4.69\% over the strongest baseline, GRPO. Similar trends hold on ImageNet-A, where accuracy rises from 37.37\% to 44.61\% and forgetting drops from 28.88\% to 20.16\%. RaPO also delivers clear gains on the more challenging TinyImageNet and CUB-200 benchmarks. In contrast, classical continual learning methods such as L2, EWC, and LwF provide only marginal or even detrimental benefits, while GRPO already serves as a strong RFT baseline that RaPO further improves by a large margin.

\vspace{-10pt}
\subsection{Class-Incremental Object Detection}
\label{sec:cil_det}
\vspace{-5pt}
\textbf{Setup.}
We further evaluate RaPO on class-incremental object detection on the COCO 2017 dataset~\cite{lin2014microsoft}, which contains 80 object categories. We design two task splits: a 5-task setting with 16 classes per task and a 10-task setting with 8 classes per task. Since an image may contain instances from multiple categories, we enforce a strict task partition: each training image is assigned to only one task, and there is no training sample overlap across tasks. For each class, at most five training images are selected, resulting in a few-shot detection setting. On the other hand, different from image classification, object detection requires fine-grained spatial perception and structured output generation. Hence, we use Qwen2-VL-7B~\cite{wang2024qwen2} as the base MLLM. We train each task for 5 epochs so that the model can sufficiently adapt to the detection objective while avoiding overfitting. For evaluation, we follow the metric definitions in Section~\ref{sec:cil_cls}, while replacing classification accuracy with box average precision (AP) under the standard COCO protocol~\cite{lin2014microsoft,lin2017feature}. Accordingly, we report last box AP $\mathcal{A}_b$ and box forgetting $\mathcal{F}_b$.

\textbf{Results.}
As listed in Table~\ref{tab:det_cl}, RaPO achieves the best $\mathcal{A}_b$ and the lowest $\mathcal{F}_b$ in both task settings. Compared with GRPO, it improves $\mathcal{A}_b$ from 14.64\% to 19.31\% and reduces $\mathcal{F}_b$ from 6.67\% to 1.39\% in the 5-task setting, and further improves $\mathcal{A}_b$ from 14.30\% to 19.12\% while reducing $\mathcal{F}_b$ from 6.73\% to 1.37\% in the 10-task setting. Unlike in image classification, LwF becomes competitive with GRPO in object detection, suggesting that distilling predictions from the preceding-task model is particularly useful for preserving structured localization behavior. Nevertheless, RaPO remains clearly superior to all baselines.

\begin{table}[t]
  \centering
  \caption{
A comparison of different methods on class-incremental video classification.
  }
  \label{tab:video_cl}
  \vspace{-0.5em}
  \resizebox{0.95\textwidth}{!}{
    \begin{tabular}{l cccccccc}
    \toprule
    \multirow{3}[2]{*}{\textbf{Method}} & \multicolumn{4}{c}{\textbf{UCF-101}} & \multicolumn{4}{c}{\textbf{Kinetics-200}} \\
    \cmidrule(lr){2-5} \cmidrule(lr){6-9}
    & \multicolumn{2}{c}{\textbf{5 Tasks}} & \multicolumn{2}{c}{\textbf{10 Tasks}} & \multicolumn{2}{c}{\textbf{5 Tasks}} & \multicolumn{2}{c}{\textbf{10 Tasks}} \\
    \cmidrule(lr){2-3} \cmidrule(lr){4-5} \cmidrule(lr){6-7} \cmidrule(lr){8-9}
    & $\mathcal{A}\uparrow$ & $\mathcal{F}\downarrow$ & $\mathcal{A}\uparrow$ & $\mathcal{F}\downarrow$ & $\mathcal{A}\uparrow$ & $\mathcal{F}\downarrow$ & $\mathcal{A}\uparrow$ & $\mathcal{F}\downarrow$ \\
    \midrule
    \multicolumn{9}{c}{\textbf{Joint Training}} \\
    SFT & 79.54 & - & 79.54 & - & 75.41 & - & 75.41 & - \\
    GRPO & 81.39 & - & 81.39 & - & 72.37 & - & 72.37 & - \\
    \midrule
    \multicolumn{9}{c}{\textbf{Continual Learning}} \\
    SFT     & 72.57\textsubscript{$\pm$2.57} & 26.00\textsubscript{$\pm$8.07} & 62.96\textsubscript{$\pm$2.65} & 27.35\textsubscript{$\pm$4.38} & 66.29\textsubscript{$\pm$1.05} & 53.28\textsubscript{$\pm$2.61} & 33.89\textsubscript{$\pm$4.10} & 61.53\textsubscript{$\pm$3.59} \\
    L2      & 72.27\textsubscript{$\pm$2.50} & 26.65\textsubscript{$\pm$8.40} & 63.80\textsubscript{$\pm$1.73} & 25.80\textsubscript{$\pm$3.19} & 66.32\textsubscript{$\pm$1.21} & 53.95\textsubscript{$\pm$3.20} & 33.81\textsubscript{$\pm$3.85} & 61.21\textsubscript{$\pm$3.88} \\
    EWC     & 72.43\textsubscript{$\pm$2.98} & 27.01\textsubscript{$\pm$9.37} & 63.76\textsubscript{$\pm$1.96} & 25.92\textsubscript{$\pm$3.23} & 66.62\textsubscript{$\pm$0.91} & 51.59\textsubscript{$\pm$3.41} & 34.67\textsubscript{$\pm$3.62} & 60.67\textsubscript{$\pm$2.90} \\
    LwF     & 74.07\textsubscript{$\pm$0.70} & 17.28\textsubscript{$\pm$2.57} & 65.00\textsubscript{$\pm$0.30} & 21.63\textsubscript{$\pm$0.92} & 68.56\textsubscript{$\pm$0.14} & 30.89\textsubscript{$\pm$1.20} & 42.14\textsubscript{$\pm$3.76} & 49.27\textsubscript{$\pm$3.24} \\
    GRPO    & 77.62\textsubscript{$\pm$0.52} & 13.81\textsubscript{$\pm$2.15} & 68.12\textsubscript{$\pm$1.72} & 16.94\textsubscript{$\pm$4.48} & 70.33\textsubscript{$\pm$1.92} & 30.37\textsubscript{$\pm$5.35} & 50.67\textsubscript{$\pm$2.09} & 34.65\textsubscript{$\pm$3.21} \\
    \rowcolor[rgb]{ 0.8, 1, 1}\textbf{RaPO (Ours)} & \textbf{79.57}\textsubscript{$\pm$0.77} & \textbf{11.19}\textsubscript{$\pm$2.63} & \textbf{71.79}\textsubscript{$\pm$1.84} & \textbf{10.92}\textsubscript{$\pm$0.15} & \textbf{74.18}\textsubscript{$\pm$2.36} & \textbf{16.76}\textsubscript{$\pm$3.73} & \textbf{54.85}\textsubscript{$\pm$2.81} & \textbf{27.28}\textsubscript{$\pm$1.38} \\
    \bottomrule
    \end{tabular}%
  }
\vspace{-18.5pt}
\end{table}

\vspace{-8pt}
\subsection{Class-Incremental Video Classification}
\label{sec:cil_video}
\vspace{-8pt}
\textbf{Setup.}
We extend the evaluation to class-incremental video classification, where the model sequentially learns new classes from a stream of video clips. 
The primary difference from image classification is the modality shift from images to video, which introduces temporal cues across tasks. We conduct experiments on two commonly used video recognition datasets: UCF-101~\cite{soomro2012ucf101} and Kinetics-200~\cite{kay2017kinetics,xie2018rethinking}. 
Due to the substantial computational cost of video training and limited hardware resources, we adopt 5-task and 10-task settings for both datasets. All other experimental settings follow the configuration described in Section~\ref{sec:cil_cls}.
\par
\textbf{Results.}
As listed in Table~\ref{tab:video_cl}, our method achieves leading performance on both datasets. On UCF-101 under the 10-task setting, RaPO attains an $\mathcal{A}$ of 71.79\% with only 10.92\% $\mathcal{F}$, compared to 68.12\% and 16.94\% for GRPO. 
On Kinetics-200 under the 5-task setting, RaPO improves $\mathcal{A}$ from 70.33\% to 74.18\% and reduces $\mathcal{F}$ from 30.37\% to 16.76\%. 
These results demonstrate that RaPO generalizes effectively to the video domain, preserving both spatial and temporal knowledge throughout the incremental learning process.

\begin{table}[t]
  \centering
  \caption{A comparison of Domain-Incremental Learning (DIL) across image classification and object detection.}
  \label{tab:dil_all}
  \vspace{-0.5em}
  \resizebox{0.75\textwidth}{!}{
    \begin{tabular}{l cccccc}
    \toprule
    \multirow{3}[2]{*}{\textbf{Method}} & \multicolumn{4}{c}{\textbf{Image Classification}} & \multicolumn{2}{c}{\textbf{Object Detection}} \\
    \cmidrule(lr){2-5} \cmidrule(lr){6-7}
    & \multicolumn{2}{c}{\textbf{DomainNet (6 Domains)}} & \multicolumn{2}{c}{\textbf{OfficeHome (4 Domains)}} & \multicolumn{2}{c}{\textbf{Pascal Series (4 Domains)}} \\
    \cmidrule(lr){2-3} \cmidrule(lr){4-5} \cmidrule(lr){6-7}
    & $\mathcal{A}\uparrow$ & $\mathcal{F}\downarrow$ & $\mathcal{A}\uparrow$ & $\mathcal{F}\downarrow$ & $\mathcal{\bar{A}}_b\uparrow$ & $\mathcal{F}_b\downarrow$ \\
    \midrule
    \multicolumn{7}{c}{\textbf{Joint Training}} \\
    SFT & 67.76 & - & 93.84 & - & 35.40 & - \\
    GRPO & 68.46 & - & 94.65 & - & 39.13 & - \\
    \midrule
    \multicolumn{7}{c}{\textbf{Continual Learning}} \\
    SFT     & 63.09\textsubscript{$\pm$0.21} & 1.34\textsubscript{$\pm$0.33} & 91.07\textsubscript{$\pm$1.13} & 1.02\textsubscript{$\pm$1.02} & 34.55\textsubscript{$\pm$2.16} & \textbf{0.55}\textsubscript{$\pm$0.73} \\
    L2      & 63.90\textsubscript{$\pm$1.55} & 2.36\textsubscript{$\pm$1.49} & 91.62\textsubscript{$\pm$0.63} & 2.29\textsubscript{$\pm$0.67} & 33.45\textsubscript{$\pm$2.98} & 1.07\textsubscript{$\pm$1.14} \\
    EWC     & 63.08\textsubscript{$\pm$1.60} & 1.28\textsubscript{$\pm$1.33} & 91.29\textsubscript{$\pm$1.19} & 2.52\textsubscript{$\pm$0.86} & 33.52\textsubscript{$\pm$2.21} & 0.61\textsubscript{$\pm$0.70} \\
    LwF     & 62.63\textsubscript{$\pm$4.15} & 1.75\textsubscript{$\pm$1.90} & 88.33\textsubscript{$\pm$1.40} & 2.82\textsubscript{$\pm$1.55} & 33.86\textsubscript{$\pm$1.41} & 0.63\textsubscript{$\pm$0.81} \\
    GRPO    & 64.54\textsubscript{$\pm$1.03} & 0.31\textsubscript{$\pm$0.26} & 92.80\textsubscript{$\pm$0.28} & 0.55\textsubscript{$\pm$0.09} & 34.58\textsubscript{$\pm$1.75} & 1.18\textsubscript{$\pm$1.44} \\
    \rowcolor[rgb]{ 0.8, 1, 1}\textbf{RaPO (Ours)} & \textbf{66.27}\textsubscript{$\pm$0.62} & \textbf{0.17}\textsubscript{$\pm$0.27} & \textbf{93.63}\textsubscript{$\pm$0.14} & \textbf{0.45}\textsubscript{$\pm$0.49} & \textbf{37.18}\textsubscript{$\pm$1.44} & 0.84\textsubscript{$\pm$0.62} \\
    \bottomrule
    \end{tabular}%
  }
  \vspace{-15pt}
\end{table}

\vspace{-8pt}
\subsection{Domain-Incremental Image Classification and Object Detection}
\label{sec:dil}
\vspace{-5pt}
\textbf{Setup.}
We evaluate RaPO under domain-incremental learning (DIL), where the model sequentially adapts to new visual domains while the label space remains fixed and no data replay from previous domains is allowed. For image classification, we use DomainNet~\cite{Peng2019domainnet} with 6 domains and OfficeHome~\cite{venkateswara2017officehome} with 4 domains. All remaining settings follow Section~\ref{sec:cil_cls}. For domain-incremental object detection, we use the four Pascal Series domains~\cite{inoue2018cross}, including Pascal VOC \cite{everingham2015pascalvoc}, Clipart, Watercolor, and Comic. All four domains share 6 common object classes, ensuring a consistent label space throughout the task sequence. 
Since the four domains have test sets of vastly different sizes, a direct all-sample AP is dominated by the largest domain. 
Therefore, we compute the AP of each domain separately and then average these four values after the model finishes the last task, termed $\mathcal{\bar{A}}_b$.
All other training and evaluation protocols remain identical to those in Section~\ref{sec:cil_det}.
\par
\textbf{Results.}
Table~\ref{tab:dil_all} shows that RaPO consistently improves performance across domain-incremental classification and detection. For example, RaPO improves $\mathcal{A}$ from 64.54\% to 66.27\% and reduces $\mathcal{F}$ from 0.31\% to 0.17\% on DomainNet compared with GRPO. For domain-incremental object detection, RaPO achieves the best $\mathcal{\bar{A}}_b$ among all continual learners at 37.18\%, improving over GRPO by 2.60\%. This result is also higher than joint-training SFT at 35.40\% and remains close to joint-training GRPO at 39.13\%. Although RaPO exhibits a slightly higher $\mathcal{F}_b$ than plain SFT, the accompanying boost in absolute detection performance is substantial, confirming that RaPO learns to preserve both location and category knowledge more effectively rather than merely resisting change.
Compared with CIL, we also observe that DIL exhibits intrinsically lower forgetting. This is because DIL maintains a fixed label space across distribution shifts, which inherently avoids the severe inter-task confusion that drives catastrophic forgetting when learning disjoint novel classes in CIL \cite{van2022three}.

\vspace{-8pt}
\section{Conclusion}
\vspace{-8pt}
In this work, we investigate the potential of RFT for visual continual learning. 
Through a pilot study, we empirically confirm that RFT is more resilient to catastrophic forgetting than SFT, yet still suffers from non-negligible forgetting, due to trajectory-level drift agnosticism. 
To this end, we propose \textbf{RaPO}, which addresses catastrophic forgetting through two complementary components: a {Retention Reward} that preferentially reinforces low-drift, knowledge-preserving rollouts within each group, and CTAN that stabilizes the optimization scale across task-wise boundaries. 
Extensive results across diverse visual continual learning settings demonstrate that RaPO consistently achieves leading performance. 
As the first systematic exploration of RFT in visual continual learning, we hope this work will stimulate further research in this area.

\appendix


\section*{Appendix}
\section{Policy-Gradient Compatibility and Optimization Stability of RaPO}
\label{sec:appendix_optimization_stability}

We provide a standard reinforcement-learning view of RaPO. This analysis does not claim global optimality or a formal no-forgetting guarantee. Instead, it focuses on three optimization properties of the proposed retention reward: compatibility with the score-function policy-gradient estimator under a detached-reward surrogate view, bounded reward and advantage magnitudes, and the standard stationary-point guarantee inherited under idealized non-convex stochastic policy-gradient assumptions.

\textbf{Policy-gradient compatibility.}
At optimization step $k$, let $\theta_k$ denote the policy used to sample rollouts and compute the retention reward. For a sampled trajectory $y=(y_1,\ldots,y_m)$, define the length-normalized pre-truncation log-ratio drift as
\begin{equation}
d_{\theta_k}(y)
=
\frac{1}{m}
\sum_{s=1}^{m}
\left[
\log \pi_{\theta_k}(y_s\mid x,y_{<s})
-
\log \pi_{t-1}(y_s\mid x,y_{<s})
\right].
\label{eq:appendix_pg_drift}
\end{equation}
This quantity is the trajectory-level counterpart of the drift proxy used in Equation~\eqref{eq:rapo_dbar}. Under on-policy sampling from $\pi_{\theta_k}$, its expectation corresponds to a token-averaged forward log-ratio contribution of $\pi_{\theta_k}$ relative to the anchor $\pi_{t-1}$, whereas a single sampled value can be negative and should not be interpreted as a non-negative KL value. RaPO therefore uses the one-sided truncation $[d_{\theta_k}(y)]_+$ inside the reward, but does not optimize this signal as an explicit differentiable regularizer. As in standard PPO/GRPO implementations, the reward value assigned to each sampled rollout is treated as stop-gradient scalar feedback when forming the policy-gradient update. Concretely, after rollouts are collected at $\theta_k$, RaPO constructs the detached shaped reward
\begin{equation}
\widetilde{R}_k(y)
=
R_{\mathrm{task}}(y)
+
\lambda
\exp
\left(
-\alpha
\left[
d_{\theta_k}(y)
\right]_+
\right),
\label{eq:appendix_detached_reward}
\end{equation}
where the subscript $k$ emphasizes that the scalar reward is fixed during the subsequent gradient computation. This distinction is important: if the log-ratio inside $R_{\mathrm{ret}}$ were differentiated directly, it would become a loss-level KL-style regularizer. RaPO instead uses the drift only to shape the sampled rollout's reward, so the policy-gradient term remains a likelihood-ratio update weighted by a scalar return.
\par
To make this precise, consider the prompt-conditioned local detached objective
\begin{equation}
\mathcal{J}_k(\theta;x)
=
\mathbb{E}_{y\sim\pi_{\theta}(\cdot\mid x)}
\left[
\widetilde{R}_k(y)
\right].
\label{eq:appendix_local_objective}
\end{equation}
This objective is an analysis device for a single update: $\widetilde{R}_k$ is fixed while differentiating with respect to $\theta$, although the detached reward rule is recomputed after new rollouts are collected at later optimization steps.
Refer to~\cite{liu2025drgrpo}, for any baseline $b(x)$ that does not depend on the sampled trajectory $y$, the score-function identity gives
\begin{equation}
\mathbb{E}_{y\sim\pi_{\theta_k}(\cdot\mid x)}
\left[
\left(
\widetilde{R}_k(y)-b(x)
\right)
\nabla_\theta
\log
\pi_{\theta}(y\mid x)
\bigg|_{\theta=\theta_k}
\right]
=
\nabla_\theta
\mathcal{J}_k(\theta;x)
\bigg|_{\theta=\theta_k}.
\label{eq:appendix_unbiased_pg}
\end{equation}
The equality follows because $\widetilde{R}_k(y)$ is detached: the gradient acts only on $\log \pi_{\theta}(y\mid x)$, while the reward is treated exactly like a verifier-provided return. Therefore, in the ideal score-function form, adding the retention reward does not introduce an additional pathwise gradient term; it simply changes the scalar credit assigned to each sampled trajectory. In practical GRPO, clipping and finite-group normalization make the estimator an approximation to this ideal identity, and the group mean is a sample-dependent baseline rather than a fixed $b(x)$. The key point is narrower: the retention term itself does not add a separate token-level distillation gradient that directly pulls every token distribution toward the anchor.

\textbf{Bounded reward and advantage.}
Assume the task verifier is bounded, i.e.,
\begin{equation}
0 \leq R_{\mathrm{task}}(y) \leq R_{\max}.
\end{equation}
Since the retention reward is defined as
\begin{equation}
R_{\mathrm{ret}}(y)=\exp(-\alpha \bar{D}_{\mathrm{drift}}(y)),
\end{equation}
and $\bar{D}_{\mathrm{drift}}(y)\geq 0$, we have
\begin{equation}
0 < R_{\mathrm{ret}}(y) \leq 1.
\end{equation}
Therefore, the total reward used by RaPO is bounded as
\begin{equation}
0 \leq R_{\mathrm{total}}(y)
=R_{\mathrm{task}}(y)+\lambda R_{\mathrm{ret}}(y)
\leq R_{\max}+\lambda.
\label{eq:appendix_total_reward_bound}
\end{equation}
Let $B=R_{\max}+\lambda$. For a rollout group, the group mean reward also lies in $[0,B]$, and CTAN computes the advantage with denominator $\hat{\sigma}+\epsilon$. Since $\hat{\sigma}$ is an EMA of non-negative batch standard deviations, $\hat{\sigma}\geq 0$ whenever it is initialized non-negatively. Thus,
\begin{equation}
|A_i^{\mathrm{RaPO}}|
=
\left|
\frac{R_{\mathrm{total}}(y_i)-\mu_{\mathrm{group}}}{\hat{\sigma}+\epsilon}
\right|
\leq
\frac{B}{\epsilon}.
\label{eq:appendix_advantage_bound}
\end{equation}
This shows that the retention reward can enlarge the reward range by at most $\lambda$, while CTAN keeps the normalization denominator lower-bounded by $\epsilon$. Consequently, RaPO does not introduce unbounded advantages. To connect this bound to the usual bounded-gradient condition, we additionally require the standard score-function moment assumption: the trajectory score $\nabla_\theta \log \pi_\theta(y\mid x)=\sum_s \nabla_\theta \log \pi_\theta(y_s\mid x,y_{<s})$ has bounded second moment. Under this assumption, the score-function estimator weighted by $A_i^{\mathrm{RaPO}}$ also has a bounded second moment.

\textbf{Idealized convergence to a stationary point and proof.}
The practical RaPO update uses clipped GRPO, finite rollout groups, and a detached reward that is recomputed from the current sampling policy. A complete convergence theorem for this exact time-varying procedure would need to track the clipping bias and sample-dependent group normalization. Here we state the standard idealized stochastic policy-gradient result that RaPO inherits once its shaped reward is viewed as a bounded detached scalar feedback signal.
\par
Let $\mathcal{J}_{\mathrm{RaPO}}(\theta)$ denote a fixed detached shaped surrogate objective over the analysis window, where verifier rewards and retention rewards are treated as stop-gradient scalar feedback when forming the score-function estimator. Suppose that: 

(i) $\mathcal{J}_{\mathrm{RaPO}}$ is $L$-smooth and upper bounded by $\mathcal{J}^{\star}$; 

(ii) the stochastic policy-gradient estimator $g_k$ is unbiased for this surrogate, i.e., $\mathbb{E}[g_k\mid \theta_k]=\nabla \mathcal{J}_{\mathrm{RaPO}}(\theta_k)$; 

(iii) its variance is bounded, i.e., $\mathbb{E}[\|g_k-\nabla \mathcal{J}_{\mathrm{RaPO}}(\theta_k)\|^2\mid \theta_k]\leq \sigma_g^2$. 
These assumptions are standard in non-convex stochastic policy-gradient analyses. Assumption (i) is a local smoothness and bounded-objective condition on the detached surrogate. Assumption (ii) follows from the score-function identity in Equation~\eqref{eq:appendix_unbiased_pg} before practical approximations such as clipping and finite-group normalization are applied. Assumption (iii) is supported by the bounded reward-and-advantage result above: if the trajectory score has a bounded second moment, i.e.,
\begin{equation}
\mathbb{E}
\left[
\left\|
\nabla_\theta \log \pi_\theta(y\mid x)
\right\|^2
\mid \theta
\right]
\leq
C_{\mathrm{score}},
\end{equation}
Then, Equation~\eqref{eq:appendix_advantage_bound} implies that the idealized score-function estimator has a bounded second moment,
\begin{equation}
\mathbb{E}
\left[
\left\|
A^{\mathrm{RaPO}}
\nabla_\theta \log \pi_\theta(y\mid x)
\right\|^2
\mid \theta
\right]
\leq
\left(
\frac{B}{\epsilon}
\right)^2
C_{\mathrm{score}},
\end{equation}
which in turn yields a bounded variance term $\sigma_g^2$.

Then stochastic gradient ascent with $\theta_{k+1}=\theta_k+\eta g_k$ and $\eta\leq 1/L$ satisfies
\begin{equation}
\min_{0\leq k<K}
\mathbb{E}
\left[
\left\|
\nabla \mathcal{J}_{\mathrm{RaPO}}(\theta_k)
\right\|^2
\right]
\leq
\frac{2(\mathcal{J}^{\star}-\mathcal{J}_{\mathrm{RaPO}}(\theta_0))}{\eta K}
+
L\eta\sigma_g^2.
\label{eq:appendix_convergence_rate}
\end{equation}
Choosing $\eta=\mathcal{O}(K^{-1/2})$ yields
\begin{equation}
\min_{0\leq k<K}
\mathbb{E}
\left[
\left\|
\nabla \mathcal{J}_{\mathrm{RaPO}}(\theta_k)
\right\|^2
\right]
=
\mathcal{O}(K^{-1/2}).
\end{equation}

To prove this statement, use the smoothness of $\mathcal{J}_{\mathrm{RaPO}}$. One step of stochastic gradient ascent gives
\begin{equation}
\mathcal{J}_{\mathrm{RaPO}}(\theta_{k+1})
\geq
\mathcal{J}_{\mathrm{RaPO}}(\theta_k)
+
\eta
\left\langle
\nabla \mathcal{J}_{\mathrm{RaPO}}(\theta_k), g_k
\right\rangle
-
\frac{L\eta^2}{2}\|g_k\|^2.
\end{equation}
Taking the conditional expectation and using the unbiasedness assumption,
\begin{equation}
\mathbb{E}
\left[
\mathcal{J}_{\mathrm{RaPO}}(\theta_{k+1})
-
\mathcal{J}_{\mathrm{RaPO}}(\theta_k)
\mid \theta_k
\right]
\geq
\eta
\left\|
\nabla \mathcal{J}_{\mathrm{RaPO}}(\theta_k)
\right\|^2
-
\frac{L\eta^2}{2}
\left(
\left\|
\nabla \mathcal{J}_{\mathrm{RaPO}}(\theta_k)
\right\|^2
+
\sigma_g^2
\right).
\end{equation}
When $\eta\leq 1/L$, this implies
\begin{equation}
\mathbb{E}
\left[
\mathcal{J}_{\mathrm{RaPO}}(\theta_{k+1})
-
\mathcal{J}_{\mathrm{RaPO}}(\theta_k)
\right]
\geq
\frac{\eta}{2}
\mathbb{E}
\left[
\left\|
\nabla \mathcal{J}_{\mathrm{RaPO}}(\theta_k)
\right\|^2
\right]
-
\frac{L\eta^2}{2}\sigma_g^2.
\end{equation}
Summing the above inequality from $k=0$ to $K-1$ and using the upper bound $\mathcal{J}^{\star}$ gives Equation~\eqref{eq:appendix_convergence_rate}.

\section{More Detailed Implementations}
\label{sec:append_implementations}
\subsection{Prompt Template and Output Format}
\label{sec:appendix_prompt_templates}

\begin{figure}[t]
  \centering
  \includegraphics[width=0.96\textwidth]{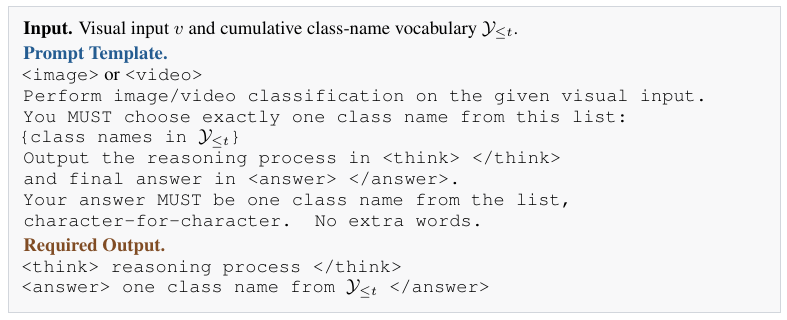}
  \caption{Prompt template and required output format for image and video classification.}
  \label{fig:prompt_cls}
\end{figure}

\textbf{Image and Video Classification.}
As shown in Figure~\ref{fig:prompt_cls}, at task $\mathcal{T}_t$, let $\Delta\mathcal{Y}_t$ denote the set of novel class names introduced by the current classification task, and let $\mathcal{Y}_{\leq t} = \mathcal{Y}_{\leq t-1} \cup \Delta\mathcal{Y}_t$ be the cumulative vocabulary of all class names observed so far. 
The prompt explicitly lists every class in $\mathcal{Y}_{\leq t}$ as the candidate answer space and never exposes future-task class names, so the model is evaluated strictly on its ability to recognize and retain classes that have already been introduced. 
The final answer is constrained to a closed-set prediction over this predefined vocabulary, while the reasoning span remains free-form and unconstrained. 
Video classification follows the same template, with the image placeholder replaced by a video placeholder. Note that in domain-incremental learning, the class vocabulary remains unchanged across all tasks since only the visual domain varies. 
Although we acknowledge the practical importance of open-set recognition in real-world applications, our primary goal is to investigate the potential of RFT in visual continual learning, where the central challenge lies in balancing plasticity to newly introduced classes against the stability of previously acquired ones. 
Allowing fully open-set generation may introduce confounding factors such as synonyms, hypernyms, or explanatory paraphrases that fall outside the evaluation taxonomy, making it difficult to isolate whether performance changes stem from genuine recognition gains or merely from naming variability.

\begin{figure}[t]
  \centering
  \includegraphics[width=0.96\textwidth]{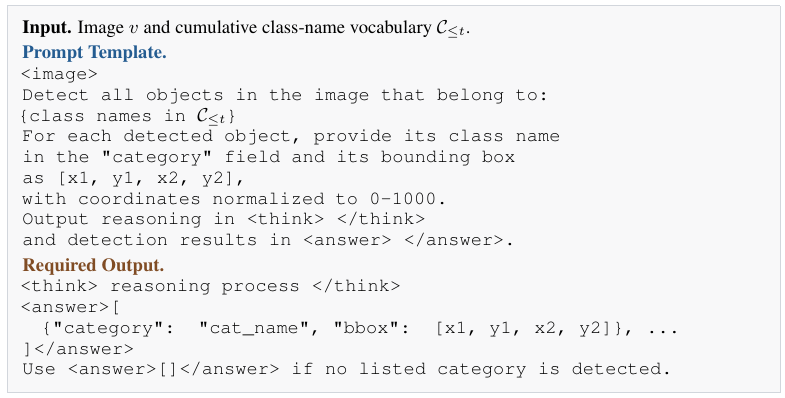}
  \caption{Prompt template and required output format for object detection.}
  \label{fig:prompt_det}
\end{figure}

\textbf{Object Detection.}
For object detection, let $\Delta\mathcal{C}_t$ denote the novel class names introduced by the current task, and let $\mathcal{C}_{\leq t} = \mathcal{C}_{\leq t-1} \cup \Delta\mathcal{C}_t$ be the cumulative seen class vocabulary up to task $\mathcal{T}_t$. 
As shown in Figure \ref{fig:prompt_det}, the model is required to output a JSON list of detected instances, where the \texttt{"category"} field of each entry is restricted to a closed-set prediction over $\mathcal{C}_{\leq t}$. 
This constraint ensures that the evaluation faithfully reflects the model's ability to recognize and retain previously learned object classes, consistent with the closed-set protocol adopted in classification. 
At the same time, the model freely predicts the number of instances and their spatial locations. 
Bounding box coordinates are expressed as integer values normalized to the range $[0, 1000]$, making the verifier independent of the original image resolutions.

\subsection{Task Reward Function Design}
\label{sec:appendix_task_rewards}
For the reward definitions below, $y_i$ denotes the $i$-th response sampled for visual input $v$ within a rollout group.

\textbf{Classification Reward.}
For image and video classification, the task reward combines answer correctness and format compliance:
\begin{equation}
R_{\mathrm{task}}(y_i) = R_{\mathrm{acc}}(y_i) + R_{\mathrm{fmt}}(y_i).
\end{equation}
Here, $R_{\mathrm{acc}} \in \{0,1\}$ is an exact-match reward for the final class name extracted from the \texttt{<answer>} span. 
The prediction and the ground-truth class name are normalized by lower-casing and by mapping underscores, hyphens, and periods to spaces before matching, making the verifier robust to superficial formatting variation while preserving class identity. 
The format term $R_{\mathrm{fmt}} \in \{0,1\}$ checks the complete \texttt{<think></think> <answer></answer>} structure. 

\textbf{Detection Reward.}
For object detection, the task reward combines localization quality, class-name accuracy, and format compliance:
\begin{equation}
R_{\mathrm{task}}(y_i) = R_{\mathrm{iou}}(y_i) + R_{\mathrm{cls}}(y_i) + R_{\mathrm{fmt}}(y_i).
\end{equation}
The verifier first parses the JSON output from the \texttt{<answer>} span into a set of predicted boxes. Class names are lower-cased, coordinates are rounded to the nearest integer and clipped to $[0, 1000]$, and degenerate boxes (e.g., those with zero area) are discarded. Following DETR~\cite{carion2020detr}, we construct a bipartite matching score matrix where same-class pairs are scored by their IoU and mismatched pairs receive zero score. The Hungarian algorithm then finds an optimal one-to-one assignment between predictions and ground truth. This matching is critical for meaningful reward computation: it ensures each ground-truth instance is associated with at most one prediction, preventing the policy from inflating its reward by outputting multiple near-identical boxes for the same object.
Based on the matching, $R_{\mathrm{iou}}$ is the sum of IoU values from all assigned same-class pairs divided by $\max(N_{\mathrm{gt}}, N_{\mathrm{pred}})$, where $N_{\mathrm{gt}}$ and $N_{\mathrm{pred}}$ denote the number of ground-truth and predicted instances, respectively. This denominator normalizes the reward by the larger of the two counts, penalizing both missed detections and excessive false positives in a single term. $R_{\mathrm{cls}}$ is the fraction of assigned same-class pairs whose IoU meets or exceeds $0.5$, computed under the same denominator. The format term $R_{\mathrm{fmt}}$ verifies both the output tag structure and JSON schema. Outputs with valid tags but unparsable JSON receive partial format credit, while $R_{\mathrm{iou}}$ and $R_{\mathrm{cls}}$ are set to zero.


\section{Ablation Studies}
\label{sec:more_ablation}
We provide a comprehensive ablation analysis of different components and hyperparameters in RaPO. Experiments are conducted on class-incremental image classification using the 10-task setting with ImageNet-R, and on class-incremental object detection using the 5-task setting with COCO 2017. 
All other experimental settings follow the configurations described in Sections \ref{sec:cil_cls} and \ref{sec:cil_det}. Unless otherwise specified, RaPO uses the default hyperparameters: 
rollout group size $n=8$ (Section \ref{sec:prelim}), retention reward scale $\alpha=20$ (Section \ref{sec:rapo_ret}), retention reward weight $\lambda=0.5$ (Section \ref{sec:rapo_ret}), and smoothing coefficient $\beta=0.999$ (Section \ref{sec:rapo_can}).

\begin{figure}[t]
    \centering
    \includegraphics[width=0.85\textwidth]{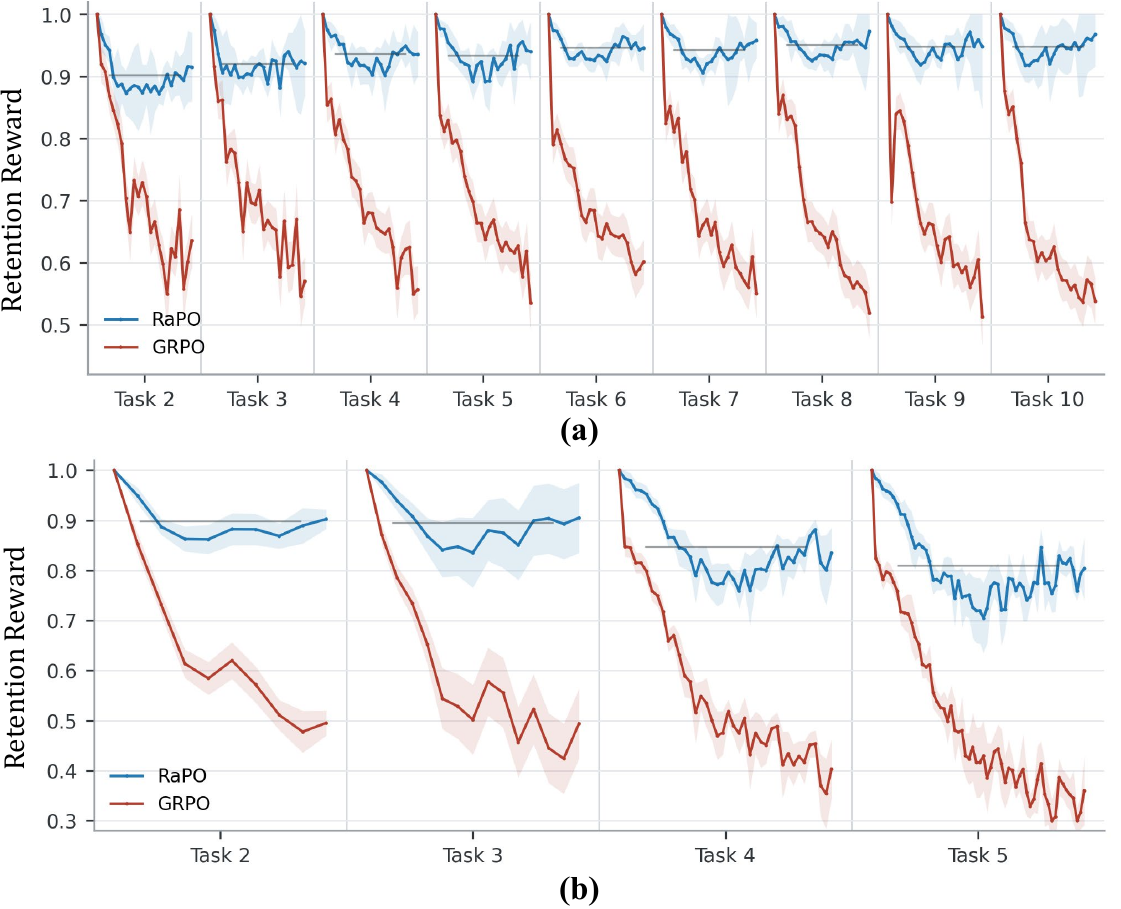}
    \caption{
    Retention reward dynamics on (a) ImageNet-R 10-task class-incremental classification and (b) COCO 5-task class-incremental detection. 
    }
    \label{fig:ret_re_curve}
\end{figure}

\subsection{Analysis of Core Components in RaPO}
\label{sec:appendix_ablation_components}
We first verify the contribution of the two core components, Retention Reward (Section~\ref{sec:rapo_ret}) and CTAN (Section~\ref{sec:rapo_can}), by gradually adding them to the vanilla GRPO baseline. As shown in Table~\ref{tab:ablation_components}, both components individually improve performance on the two benchmarks. Adding CTAN alone raises $\mathcal{A}$ from 74.67 to 79.43 and $\mathcal{A}_b$ from 14.64 to 16.27, while reducing $\mathcal{F}$ from 20.02 to 13.17 and $\mathcal{F}_b$ from 6.67 to 4.73. This demonstrates that stabilizing the advantage scale across task boundaries already benefits both task adaptation and knowledge retention. Incorporating the Retention Reward provides a substantially larger gain, pushing $\mathcal{A}$ to 82.57 and $\mathcal{A}_b$ to 18.11, with forgetting rates dropping to 8.38 and 2.61, respectively. Combining both components yields the best results across all metrics, achieving $\mathcal{A}$ of 85.92 and $\mathcal{F}$ of 4.69 on ImageNet-R, and $\mathcal{A}_b$ of 19.31 and $\mathcal{F}_b$ of 1.39 on COCO. These results confirm that CTAN and the Retention Reward play complementary roles: CTAN stabilizes the update magnitude across tasks while the Retention Reward steers credit assignment toward low-drift trajectories that preserve previously acquired knowledge.
\par
On the other hand, we further analyze the retention-reward dynamics. For RaPO, \(R_{\mathrm{ret}}\) is optimized with the default coefficient \(\lambda=0.5\). 
For a controlled diagnostic comparison, we also attach the same retention-reward estimator to vanilla GRPO while fixing its coefficient to \(\lambda=0\). Therefore, the computed \(R_{\mathrm{ret}}\) for GRPO is used only for measurement: it is not used in the advantage computation, and never contributes to policy updates. 
As shown in Figure~\ref{fig:ret_re_curve}, RaPO consistently maintains higher retention rewards than GRPO after task transitions on both datasets. 
Specifically, the GRPO curves reveal that task-reward-only optimization can move toward high-drift trajectories, even though it adapts to the current task. 
In contrast, RaPO preserves higher \(R_{\mathrm{ret}}\) throughout the continual sequence, showing that the retention reward continuously reshapes credit assignment toward low-drift, knowledge-preserving trajectories. 

\begin{table}[t]
  \centering
  \caption{Ablation study of core components in RaPO.}
  \label{tab:ablation_components}
  \vspace{-0.5em}
  \resizebox{0.75\textwidth}{!}{%
    \begin{tabular}{l cccc}
    \toprule
    \multirow{2}[2]{*}{\textbf{Method}} & \multicolumn{2}{c}{\textbf{ImageNet-R (10 Tasks)}} & \multicolumn{2}{c}{\textbf{COCO (5 Tasks)}} \\
    \cmidrule(lr){2-3} \cmidrule(lr){4-5}
    & $\mathcal{A}\uparrow$ & $\mathcal{F}\downarrow$ & $\mathcal{A}_b\uparrow$ & $\mathcal{F}_b\downarrow$ \\
    \midrule
    GRPO & 74.67\textsubscript{$\pm$1.27} & 20.02\textsubscript{$\pm$4.91} & 14.64\textsubscript{$\pm$2.18} & 6.67\textsubscript{$\pm$2.71} \\
    GRPO + CTAN & 79.43\textsubscript{$\pm$1.68} & 13.17\textsubscript{$\pm$3.26} & 16.27\textsubscript{$\pm$1.63} & 4.73\textsubscript{$\pm$1.92} \\
    GRPO + Retention Reward & 82.57\textsubscript{$\pm$1.74} & 8.38\textsubscript{$\pm$2.42} & 18.11\textsubscript{$\pm$1.31} & 2.61\textsubscript{$\pm$1.37} \\
    \rowcolor[rgb]{ .906,  .902,  .902}\textbf{RaPO (Ours)} & \textbf{85.92}\textsubscript{$\pm$1.82} & \textbf{4.69}\textsubscript{$\pm$1.71} & \textbf{19.31}\textsubscript{$\pm$1.08} & \textbf{1.39}\textsubscript{$\pm$1.20} \\
    \bottomrule
    \end{tabular}%
  }
\end{table}

\begin{table}[t]
  \centering
  \caption{Effect of retention anchor strategy.}
  \label{tab:appendix_ablation_anchor}
  \vspace{-0.5em}
  \resizebox{0.675\textwidth}{!}{%
    \begin{tabular}{l cccc}
    \toprule
    \multirow{2}[2]{*}{\textbf{Anchor Policy}} & \multicolumn{2}{c}{\textbf{ImageNet-R (10 Tasks)}} & \multicolumn{2}{c}{\textbf{COCO (5 Tasks)}} \\
    \cmidrule(lr){2-3} \cmidrule(lr){4-5}
    & $\mathcal{A}\uparrow$ & $\mathcal{F}\downarrow$ & $\mathcal{A}_b\uparrow$ & $\mathcal{F}_b\downarrow$ \\
    \midrule
    \rowcolor[rgb]{ .906,  .902,  .902}\(\pi_{t-1}\) & \textbf{85.92}\textsubscript{$\pm$1.82} & \textbf{4.69}\textsubscript{$\pm$1.71} & \textbf{19.31}\textsubscript{$\pm$1.08} & \textbf{1.39}\textsubscript{$\pm$1.20} \\
    \(\pi_{ema}\) & 84.86\textsubscript{$\pm$2.01} & 5.65\textsubscript{$\pm$2.78} & 18.61\textsubscript{$\pm$1.27} & 2.11\textsubscript{$\pm$1.32} \\
    \(\pi_{1}\)  & 83.88\textsubscript{$\pm$1.93} & 7.24\textsubscript{$\pm$2.34} & 17.34\textsubscript{$\pm$1.48} & 2.82\textsubscript{$\pm$1.57} \\
    \(\pi_{p}\) & 82.68\textsubscript{$\pm$2.17} & 9.43\textsubscript{$\pm$2.66} & 16.88\textsubscript{$\pm$1.51} & 3.58\textsubscript{$\pm$1.93} \\
    \bottomrule
    \end{tabular}%
  } 
\end{table}

\begin{table}[t]
  \centering
  \caption{Hyperparameter sensitivity in RaPO.}
  \label{tab:ablation_hparams}
  \vspace{-0.5em}
  \resizebox{0.7\textwidth}{!}{%
    \begin{tabular}{l cccc}
    \toprule
    \multirow{2}[2]{*}{\textbf{Hyperparameter}} & \multicolumn{2}{c}{\textbf{ImageNet-R (10 Tasks)}} & \multicolumn{2}{c}{\textbf{COCO (5 Tasks)}} \\
    \cmidrule(lr){2-3} \cmidrule(lr){4-5}
    & $\mathcal{A}\uparrow$ & $\mathcal{F}\downarrow$ & $\mathcal{A}_b\uparrow$ & $\mathcal{F}_b\downarrow$ \\
    \midrule
    $\lambda = 0.2$ & 84.23\textsubscript{$\pm$1.64} & 7.28\textsubscript{$\pm$2.47} & 18.13\textsubscript{$\pm$1.39} & 2.48\textsubscript{$\pm$1.52} \\
    \rowcolor[rgb]{ .906,  .902,  .902}$\lambda ={0.5}$ & {85.92}\textsubscript{$\pm$1.82} & {4.69}\textsubscript{$\pm$1.71} & {19.31}\textsubscript{$\pm$1.08} & {1.39}\textsubscript{$\pm$1.20} \\
    $\lambda = 0.8$ & 85.37\textsubscript{$\pm$1.84} & 5.14\textsubscript{$\pm$1.83} & 18.78\textsubscript{$\pm$1.21} & 1.76\textsubscript{$\pm$1.29} \\
    $\lambda = 1.0$ & 84.43\textsubscript{$\pm$2.12} & 6.47\textsubscript{$\pm$2.16} & 18.09\textsubscript{$\pm$1.44} & 2.34\textsubscript{$\pm$1.58} \\
    \midrule
    \midrule
    $\alpha = 10$ & 84.73\textsubscript{$\pm$1.79} & 5.86\textsubscript{$\pm$2.11} & 18.29\textsubscript{$\pm$1.32} & 2.39\textsubscript{$\pm$1.47} \\
    \rowcolor[rgb]{ .906,  .902,  .902}$\alpha ={20}$ & {85.92}\textsubscript{$\pm$1.82} & {4.69}\textsubscript{$\pm$1.71} & {19.31}\textsubscript{$\pm$1.08} & {1.39}\textsubscript{$\pm$1.20} \\
    $\alpha = 40$ & 85.18\textsubscript{$\pm$1.97} & 5.31\textsubscript{$\pm$1.96} & 18.68\textsubscript{$\pm$1.27} & 1.92\textsubscript{$\pm$1.33} \\
    \midrule
    \midrule
    $\beta = 0.9$ & 84.58\textsubscript{$\pm$2.03} & 6.27\textsubscript{$\pm$2.36} & 18.14\textsubscript{$\pm$1.41} & 2.53\textsubscript{$\pm$1.62} \\
    $\beta = 0.99$ & 85.33\textsubscript{$\pm$1.81} & 5.27\textsubscript{$\pm$1.91} & 18.82\textsubscript{$\pm$1.22} & 1.89\textsubscript{$\pm$1.31} \\
    \rowcolor[rgb]{ .906,  .902,  .902}$\beta = {0.999}$ & {85.92}\textsubscript{$\pm$1.82} & {4.69}\textsubscript{$\pm$1.71} & {19.31}\textsubscript{$\pm$1.08} & {1.39}\textsubscript{$\pm$1.20} \\
    $\beta = 0.9999$ & 85.63\textsubscript{$\pm$1.87} & 4.91\textsubscript{$\pm$1.78} & 19.07\textsubscript{$\pm$1.12} & 1.56\textsubscript{$\pm$1.24} \\
    \bottomrule
    \end{tabular}%
  } 
\end{table}

\begin{table}[t]
  \centering
  \caption{Effect of loss-level KL regularization.}
  \label{tab:ablation_actor_kl}
  \vspace{-0.5em}
  \resizebox{0.7\textwidth}{!}{%
    \begin{tabular}{l cccc}
    \toprule
    \multirow{2}[2]{*}{\textbf{Method}} & \multicolumn{2}{c}{\textbf{ImageNet-R (10 Tasks)}} & \multicolumn{2}{c}{\textbf{COCO (5 Tasks)}} \\
    \cmidrule(lr){2-3} \cmidrule(lr){4-5}
    & $\mathcal{A}\uparrow$ & $\mathcal{F}\downarrow$ & $\mathcal{A}_b\uparrow$ & $\mathcal{F}_b\downarrow$ \\
    \midrule
    GRPO (w/o KL Loss) & 73.63\textsubscript{$\pm$1.73} & 22.87\textsubscript{$\pm$4.37} & 13.08\textsubscript{$\pm$1.66} & 8.17\textsubscript{$\pm$2.52} \\
    \rowcolor[rgb]{ .906,  .902,  .902}GRPO (w/ KL Loss) & 74.67\textsubscript{$\pm$1.27} & 20.02\textsubscript{$\pm$4.91} & 14.64\textsubscript{$\pm$2.18} & 6.67\textsubscript{$\pm$2.71} \\
    \midrule
    RaPO (w/o KL Loss) & 84.38\textsubscript{$\pm$1.86} & 6.13\textsubscript{$\pm$1.94} & 17.89\textsubscript{$\pm$1.28} & 2.72\textsubscript{$\pm$1.36} \\
    \rowcolor[rgb]{ .906,  .902,  .902}{RaPO (w/ KL Loss)} & {85.92}\textsubscript{$\pm$1.82} & {4.69}\textsubscript{$\pm$1.71} & {19.31}\textsubscript{$\pm$1.08} & {1.39}\textsubscript{$\pm$1.20} \\
    \bottomrule
    \end{tabular}%
  }
\end{table}

\begin{table}[t]
  \centering
  \caption{Ablation study on reasoning capacity and number of rollouts ($n$).}
  \label{tab:ablation_prompt_rollout}
  \vspace{-0.5em}
  \resizebox{0.75\textwidth}{!}{%
    \begin{tabular}{l cccc}
    \toprule
    \multirow{2}[2]{*}{\textbf{Method}} & \multicolumn{2}{c}{\textbf{ImageNet-R (10 Tasks)}} & \multicolumn{2}{c}{\textbf{COCO (5 Tasks)}} \\
    \cmidrule(lr){2-3} \cmidrule(lr){4-5}
    & $\mathcal{A}\uparrow$ & $\mathcal{F}\downarrow$ & $\mathcal{A}_b\uparrow$ & $\mathcal{F}_b\downarrow$ \\
    \midrule
    \rowcolor[rgb]{ .906,  .902,  .902}GRPO (w/ Reasoning) & 74.67\textsubscript{$\pm$1.27} & 20.02\textsubscript{$\pm$4.91} & 14.64\textsubscript{$\pm$2.18} & 6.67\textsubscript{$\pm$2.71} \\
    GRPO (w/o Reasoning) & 72.83\textsubscript{$\pm$1.96} & 22.14\textsubscript{$\pm$5.27} & 9.82\textsubscript{$\pm$2.21} & 13.05\textsubscript{$\pm$2.96} \\
    \midrule
    \rowcolor[rgb]{ .906,  .902,  .902}RaPO (w/ Reasoning) & 85.92\textsubscript{$\pm$1.82} & 4.69\textsubscript{$\pm$1.71} & 19.31\textsubscript{$\pm$1.08} & 1.39\textsubscript{$\pm$1.20} \\
    RaPO (w/o Reasoning) & 83.87\textsubscript{$\pm$2.04} & 6.84\textsubscript{$\pm$2.17} & 17.64\textsubscript{$\pm$1.41} & 2.46\textsubscript{$\pm$1.62} \\
    \midrule
    \midrule
    GRPO ($n=4$) & 72.93\textsubscript{$\pm$2.01} & 23.24\textsubscript{$\pm$5.42} & 13.73\textsubscript{$\pm$2.27} & 7.48\textsubscript{$\pm$2.82} \\
    GRPO ($n=6$) & 74.12\textsubscript{$\pm$1.62} & 21.07\textsubscript{$\pm$5.11} & 14.28\textsubscript{$\pm$2.16} & 6.92\textsubscript{$\pm$2.77} \\
    \rowcolor[rgb]{ .906,  .902,  .902}GRPO ($n=8$) & 74.67\textsubscript{$\pm$1.27} & 20.02\textsubscript{$\pm$4.91} & 14.64\textsubscript{$\pm$2.18} & 6.67\textsubscript{$\pm$2.71} \\
    GRPO ($n=10$) & 74.88\textsubscript{$\pm$1.59} & 19.73\textsubscript{$\pm$4.76} & 14.78\textsubscript{$\pm$2.23} & 6.48\textsubscript{$\pm$2.68} \\
    \midrule
    RaPO ($n=4$) & 83.82\textsubscript{$\pm$2.09} & 7.16\textsubscript{$\pm$2.31} & 17.83\textsubscript{$\pm$1.42} & 2.78\textsubscript{$\pm$1.67} \\
    RaPO ($n=6$) & 85.13\textsubscript{$\pm$1.86} & 5.36\textsubscript{$\pm$1.91} & 18.83\textsubscript{$\pm$1.23} & 1.82\textsubscript{$\pm$1.32} \\
    \rowcolor[rgb]{ .906,  .902,  .902}RaPO ($n=8$) & 85.92\textsubscript{$\pm$1.82} & 4.69\textsubscript{$\pm$1.71} & 19.31\textsubscript{$\pm$1.08} & 1.39\textsubscript{$\pm$1.20} \\
    RaPO ($n=10$) & 86.07\textsubscript{$\pm$1.77} & 4.53\textsubscript{$\pm$1.68} & 19.36\textsubscript{$\pm$1.11} & 1.34\textsubscript{$\pm$1.18} \\
    \bottomrule
    \end{tabular}%
  }
\end{table}

\begin{table}[t]
  \centering
  \caption{Effect of RaPO components applied to SAPO.}
  \label{tab:ablation_sapo}
  \vspace{-0.5em}
  \resizebox{0.625\textwidth}{!}{%
    \begin{tabular}{l cccc}
    \toprule
    \multirow{2}[2]{*}{\textbf{Method}} & \multicolumn{2}{c}{\textbf{ImageNet-R (10 Tasks)}} & \multicolumn{2}{c}{\textbf{COCO (5 Tasks)}} \\
    \cmidrule(lr){2-3} \cmidrule(lr){4-5}
    & $\mathcal{A}\uparrow$ & $\mathcal{F}\downarrow$ & $\mathcal{A}_b\uparrow$ & $\mathcal{F}_b\downarrow$ \\
    \midrule
    SAPO & 76.83\textsubscript{$\pm$1.94} & 18.07\textsubscript{$\pm$4.24} & 15.28\textsubscript{$\pm$1.84} & 6.13\textsubscript{$\pm$2.38} \\
    \rowcolor[rgb]{ .906,  .902,  .902}SAPO+Ours & 84.46\textsubscript{$\pm$1.72} & 6.52\textsubscript{$\pm$2.14} & 18.43\textsubscript{$\pm$1.33} & 2.07\textsubscript{$\pm$1.42} \\
    \bottomrule
    \end{tabular}%
  }
\end{table}

\begin{table}[t]
  \centering
  \caption{Performance comparison across different MLLM backbones (Qwen2-VL vs. Qwen2.5-VL). RaPO consistently mitigates forgetting and improves overall accuracy regardless of the underlying model version.}
  \label{tab:ablation_backbone}
  \vspace{-0.5em}
  \resizebox{0.625\textwidth}{!}{%
    \begin{tabular}{l cccc}
    \toprule
    \multirow{2}[2]{*}{\textbf{Method}} & \multicolumn{2}{c}{\textbf{ImageNet-R (10 Tasks)}} & \multicolumn{2}{c}{\textbf{COCO (5 Tasks)}} \\
    \cmidrule(lr){2-3} \cmidrule(lr){4-5}
    & $\mathcal{A}\uparrow$ & $\mathcal{F}\downarrow$ & $\mathcal{A}_b\uparrow$ & $\mathcal{F}_b\downarrow$ \\
    \midrule
    \multicolumn{5}{l}{{Qwen2-VL}} \\
    GRPO & 74.67\textsubscript{$\pm$1.27} & 20.02\textsubscript{$\pm$4.91} & 14.64\textsubscript{$\pm$2.18} & 6.67\textsubscript{$\pm$2.71}\\
    \rowcolor[rgb]{ .906,  .902,  .902}\textbf{RaPO (Ours)} & \textbf{85.92}\textsubscript{$\pm$1.82} & \textbf{4.69}\textsubscript{$\pm$1.71} & \textbf{19.31}\textsubscript{$\pm$1.08} & \textbf{1.39}\textsubscript{$\pm$1.20} \\
    \midrule
    \midrule
    \multicolumn{5}{l}{{Qwen2.5-VL}} \\
    GRPO & 76.41\textsubscript{$\pm$1.38} & 20.73\textsubscript{$\pm$4.82} & 16.89\textsubscript{$\pm$2.31} & 7.52\textsubscript{$\pm$2.58} \\
    \rowcolor[rgb]{ .906,  .902,  .902}\textbf{RaPO (Ours)} & \textbf{88.27}\textsubscript{$\pm$1.72} & \textbf{5.22}\textsubscript{$\pm$1.68} & \textbf{21.68}\textsubscript{$\pm$1.19} & \textbf{2.11}\textsubscript{$\pm$1.09} \\
    \bottomrule
    \end{tabular}%
  }
\end{table}

\begin{figure}[t]
    \centering
    \includegraphics[width=0.95\textwidth]{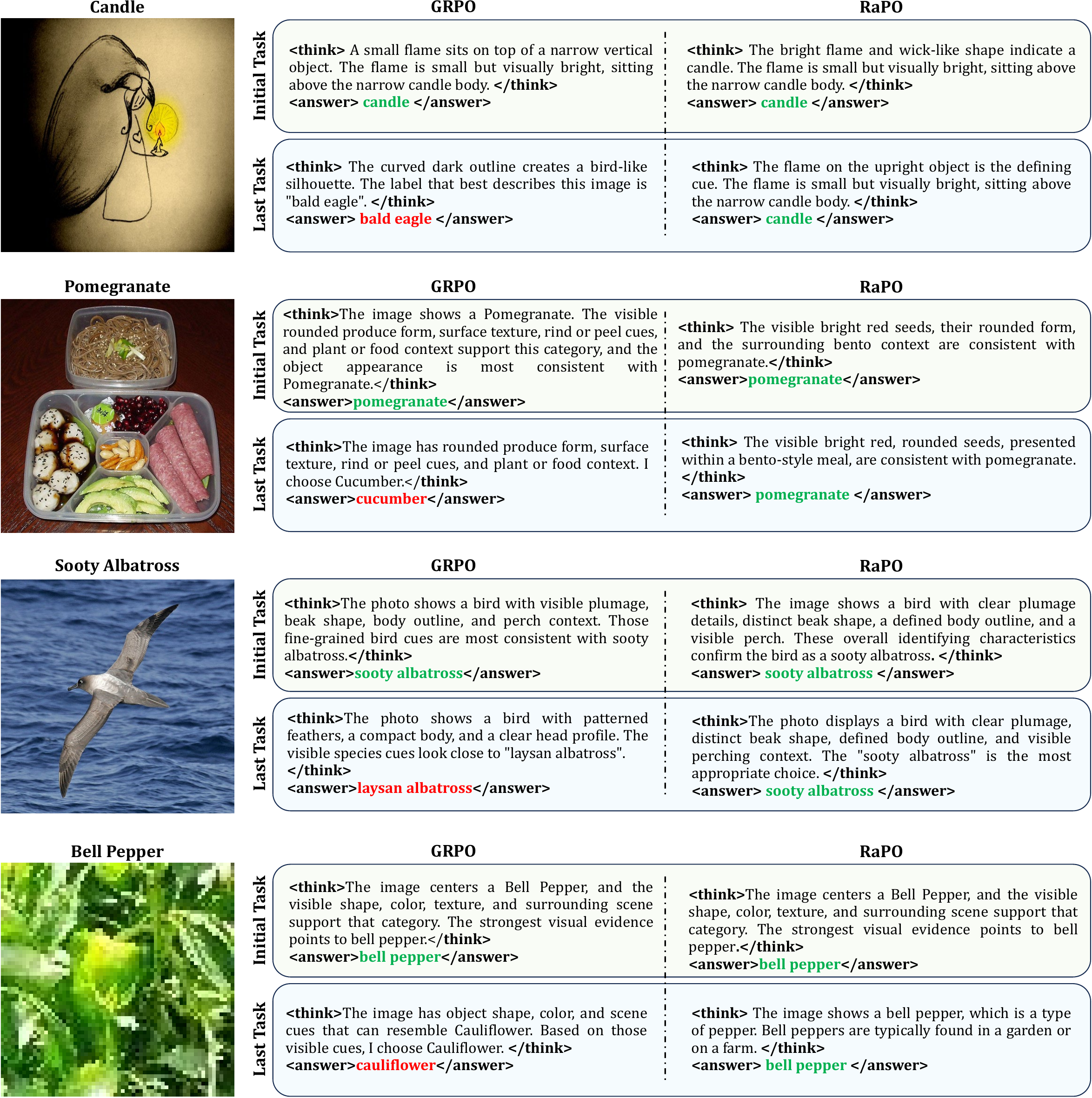}
    \caption{
    Qualitative class-incremental image classification examples from different datasets.
    }
    \label{fig:qual_cls}
\end{figure}

\begin{figure}[t]
    \centering
    \includegraphics[width=0.95\textwidth]{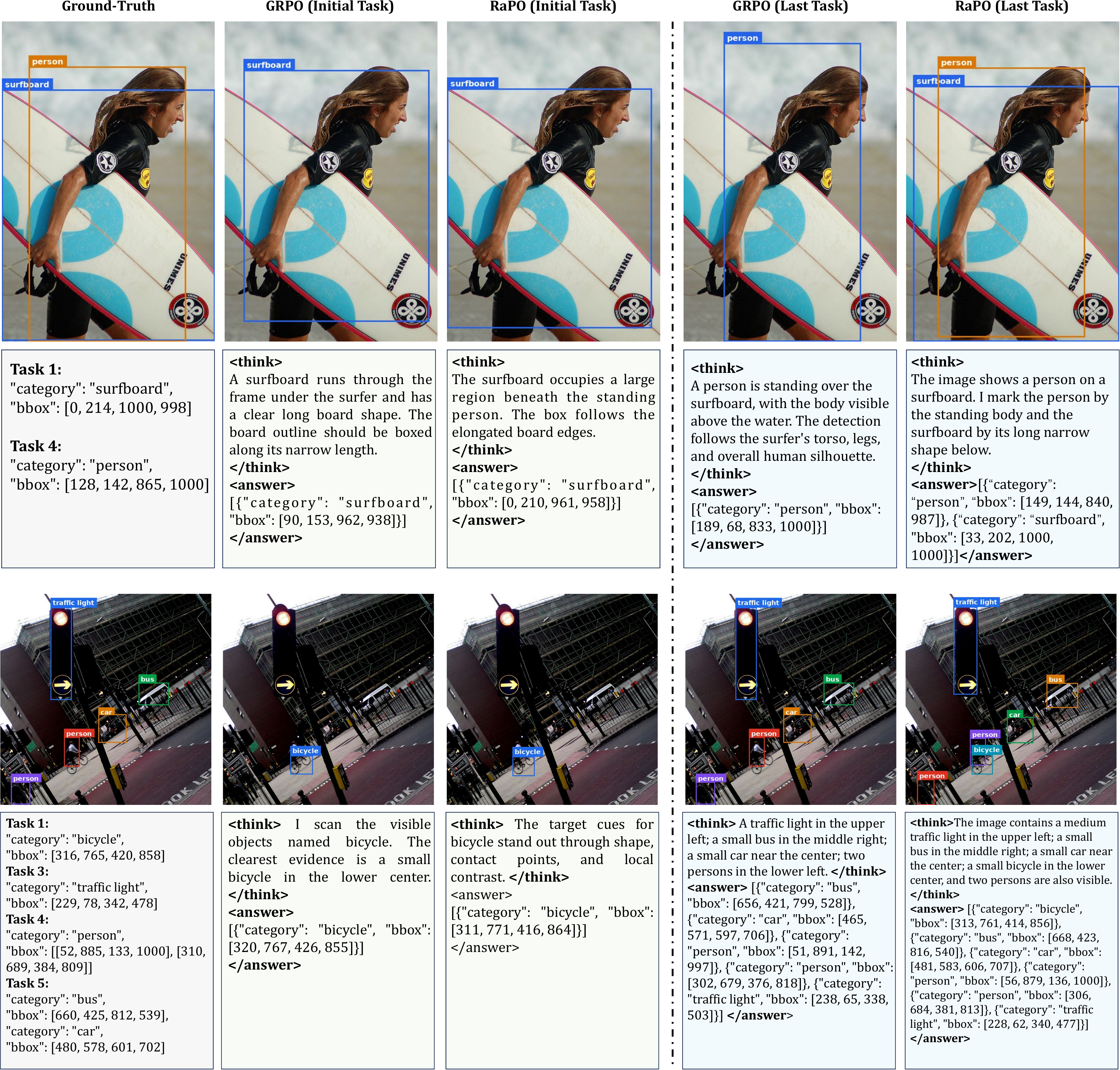}
    \caption{Qualitative class-incremental object detection examples on COCO 2017 dataset.
    }
    \label{fig:qual_det}
\end{figure}

\subsection{Effect of Anchor Policy}
We further evaluate the effect of different anchor models used for computing the retention reward, as described in Section~\ref{sec:rapo_ret}. By default, RaPO uses the immediate preceding-task policy \(\pi_{t-1}\) as the anchor. We compare this against three alternatives: an EMA of historical policies \(\pi_{ema}\) (with a decay factor of 0.1), the fixed policy from the first task \(\pi_{1}\), and the pre-trained base model \(\pi_{p}\) without any fine-tuning. As reported in Table~\ref{tab:appendix_ablation_anchor}, the preceding-task anchor achieves the best results on both classification and detection. Replacing it with the EMA anchor \(\pi_{ema}\) slightly reduces \(\mathcal{A}\) from 85.92 to 84.86 on ImageNet-R and \(\mathcal{A}_b\) from 19.31 to 18.61 on COCO. Further reverting to the task\#1 anchor \(\pi_{1}\) drops \(\mathcal{A}\) to 83.88 and \(\mathcal{A}_b\) to 17.34, with corresponding increases in forgetting. Using the pre-trained model \(\pi_{p}\) as an anchor leads to the most degradation, decreasing \(\mathcal{A}\) to 82.68 and \(\mathcal{A}_b\) to 16.88.
This trend is expected: the preceding-task policy provides a tighter and more recent reference for measuring distribution drift.
While the EMA incorporates historical information, it responds slowly to abrupt distribution shifts at task boundaries, making it less effective at constraining the large distribution changes that cause forgetting. Meanwhile, static anchors like \(\pi_{1}\) or \(\pi_{p}\) are progressively less informative about the accumulated knowledge that should be preserved as the task sequence progresses. Nevertheless, even with the weakest anchor, RaPO still outperforms vanilla GRPO by a clear margin on both benchmarks.

\subsection{Hyperparameter Sensitivity}
\label{sec:appendix_ablation_hparams}
We further examine the sensitivity of RaPO to its three key hyperparameters: the retention weight \(\lambda\), the retention sensitivity \(\alpha\), and the CTAN smoothing coefficient \(\beta\). In each experiment, only one hyperparameter is varied while the other two are held fixed. As reported in Table~\ref{tab:ablation_hparams}, varying \(\lambda\) from 0.2 to 1.0, \(\alpha\) from 10 to 40, and \(\beta\) from 0.9 to 0.9999 all produce comparable results on both benchmarks. Across all tested configurations, the last accuracy on ImageNet-R stays within a \textasciitilde 1.7\% range and the forgetting rate within a \textasciitilde 2.6\% range. Similarly, on COCO, \(\mathcal{A}_b\) varies by at most \textasciitilde 1.2\% and \(\mathcal{F}_b\) by at most \textasciitilde 1.1\%. These results indicate that RaPO is robust to hyperparameter choice and does not rely on delicate tuning.

\subsection{Impact of Standard Loss-level KL Regularization}
\label{sec:appendix_ablation_kl}
We evaluate the effect of the standard KL regularization term by removing it from both GRPO and RaPO during training. 
As shown in Table~\ref{tab:ablation_actor_kl}, removing the KL term does not lead to a large drop in performance for both methods. 
This suggests that the loss-level KL penalty does not play the primary role of stronger resistance to catastrophic forgetting in RFT, which aligns with the findings of~\cite{lai2025rlnaturally,shenfeld2026rl_razor}. 

\subsection{Effect of Reasoning and Number of Rollouts}
We further examine two practical factors: the effect of removing the reasoning step before answering, and the number of rollouts per prompt. 
For the reasoning ablation, the default prompt forces the model to reason before outputting the final answer, while the ablated version directly requests the answer without intermediate reasoning. As listed in Table~\ref{tab:ablation_prompt_rollout}, removing reasoning degrades both accuracy and forgetting for GRPO and RaPO. On class-incremental image classification, GRPO drops from 74.67 to 72.83 in $\mathcal{A}$ and RaPO from 85.92 to 83.87 in $\mathcal{A}$, a similar absolute decline of roughly 2\% for both methods. On class-incremental object detection, however, the gap is much larger for GRPO: $\mathcal{A}_b$ falls from 14.64 to 9.82, and forgetting nearly doubles. RaPO also declines on detection, but to a far lesser extent ($\mathcal{A}_b$ from 19.31 to 17.64, $\mathcal{F}_b$ from 1.39 to 2.46). 
We attribute this asymmetry to the retention reward mechanism. During training, the model naturally learns to produce implicit reasoning traces to maximize the task reward without an explicit chain-of-thought. RaPO then preferentially reinforces rollouts that remain close to the preceding-task policy, which anchors the policy to these already learned implicit reasoning patterns. Consequently, even when the prompt omits the explicit reasoning instruction, the retention signal continues to guide the policy toward trajectories that exhibit a similar expressive pattern. In contrast, GRPO's behavior is fully prompt-dependent and changes sharply when reasoning is removed.
\par
For the rollout experiments, we vary the number of sampled responses per prompt from 4 to 10. Both GRPO and RaPO improve steadily as the number of rollouts increases from 4 to 8. Beyond 8, the gains become marginal. For RaPO on ImageNet-R, moving from 8 to 10 rollouts improves $\mathcal{A}$ from 85.92 to 86.07 and $\mathcal{F}$ from 4.69 to 4.53. On COCO, $\mathcal{A}_b$ changes from 19.31 to 19.36 and $\mathcal{F}_b$ from 1.39 to 1.34. These differences are small, suggesting that 8 rollouts already provide sufficient diversity for stable advantage estimation, and further increasing the rollout count yields negligible improvement with additional computational overhead.

\section{Analytics Experiments}
\subsection{Impact of Different Policy Optimization Methods}
To examine whether RaPO is tied to GRPO, we integrate its two core components, the retention reward and CTAN, into SAPO \cite{gao2025soft}, a recent policy optimization algorithm. As shown in Table~\ref{tab:ablation_sapo}, SAPO alone yields an $\mathcal{A}$ of 76.83\% and $\mathcal{F}$ of 18.07\% on ImageNet-R, together with an $\mathcal{A}_b$ of 15.28\% and $\mathcal{F}_b$ of 6.13\% on COCO. After equipping SAPO with the retention reward and CTAN, accuracy rises to 84.46\% on classification and 18.43\% on detection, while forgetting drops to 6.52\% and 2.07\%, respectively. These improvements indicate that the retention mechanism and advantage normalization are not specific to a single optimizer. Rather, they serve as generic components that can strengthen different policy optimization algorithms for visual continual learning.

\subsection{Impact of Different MLLMs}
We assess both GRPO and RaPO on a more powerful Qwen2.5-VL model (3B for classification and 7B for detection). As shown in Table~\ref{tab:ablation_backbone}, both GRPO and RaPO gain accuracy from the stronger base models, while forgetting also increases slightly. 
This is because more capable MLLMs reach higher accuracy on early tasks, resulting in a higher performance ceiling from which any subsequent degradation is measured. Since forgetting captures the absolute drop from historical best accuracy, a stronger starting point naturally amplifies the measured forgetting for both methods.

\subsection{Qualitative Results}
\label{app:qualitative}
We present qualitative examples of class-incremental image classification on ImageNet-R, ImageNet-A, CUB-200-2011, and TinyImageNet. For each input image, we present the model response after learning the first task and the response on the same validation image after the model finishes learning the final task. As shown in Figure~\ref{fig:qual_cls}, at the initial task, both GRPO and RaPO correctly recognize the target class. After the full continual learning sequence, however, GRPO often changes its final answer to a semantically related category, which reflects forgetting of previously learned classes. 
In contrast, our RaPO preserves the correct class more consistently. The reasoning process and final answer further make this difference clear: RaPO continues to rely on the visual cues that support the old class, whereas GRPO becomes more susceptible to partial cues that lead to inaccurate decisions after more tasks are learned. 
Meanwhile, Figure~\ref{fig:qual_det} follows the same protocol for class-incremental object detection on COCO. We show the model response on the same validation image after learning the initial task that first introduces the earlier category, and again after the model finishes learning the final task. 
At the initial task, both GRPO and RaPO correctly localize the bicycle. After the full continual learning sequence, GRPO detects later categories such as traffic light, person, and car, but misses the bicycle learned earlier. This behavior reflects forgetting in continual detection: the bicycle is small and visually coupled with the rider, so its response is more easily dominated by newly introduced categories after later-task training. Conversely, RaPO still locates the bicycle while also recognizing the later categories, showing stronger retention of earlier knowledge throughout continual learning.

\section{Limitations}
\label{sec:limit}
Although RaPO demonstrates broad effectiveness across diverse visual continual learning settings, several limitations suggest promising directions for future work. First, RFT-based MLLM training is computationally expensive, and our hardware budget limits us from scaling to longer task sequences. Second, this study focuses on CIL and DIL with Qwen2-VL at the 2B and 7B scales, leaving open whether RaPO extends to larger MLLMs and more complex task formulations. Third, to isolate the properties of RFT in visual continual learning, we adopt closed-world settings with explicit task boundaries and verifiable rewards. Extending RaPO to open-world streams without such boundaries is an important direction for future work.
However, these limitations do not weaken the central contribution: RaPO is the first systematic study of RFT for visual continual learning and offers a simple yet effective strategy that consistently suppresses catastrophic forgetting phenomenon across diverse settings.


{
\small
\bibliography{references}
\bibliographystyle{unsrt}
}




\end{document}